%% file: main.tex
\documentclass{article}

     \PassOptionsToPackage{numbers}{natbib}
\usepackage[preprint]{neurips_2023}

\usepackage{hyperref}       % hyperlinks
\usepackage{url}            % simple URL typesetting
\usepackage{booktabs}       % professional-quality tables
\usepackage{amsfonts}       % blackboard math symbols
\usepackage{nicefrac}       % compact symbols for 1/2, etc.
\usepackage{microtype}      % microtypography
\usepackage[dvipsnames]{xcolor}         % colors

\usepackage{comment}
\usepackage{color}
\usepackage{adjustbox}
\usepackage{blindtext}

\usepackage{multicol}
\usepackage{multirow}
\usepackage{amsmath}
\usepackage{mathtools}
\usepackage{amssymb}
\usepackage{comment}

\usepackage{caption}
\usepackage{subcaption}

\usepackage{wrapfig}
\usepackage{bbding}
\usepackage{pifont}

\newcommand{\perf}[2]{#1\scriptsize{$\pm$ #2}}

\usepackage{enumitem}
\usepackage{algorithm}
\usepackage{algorithmicx}
\usepackage{algpseudocode} 
\algnewcommand\And{\textbf{and}}

\title{Magnitude Attention-based Dynamic Pruning}

\author{%
  Jihye Back$^{1}$ \qquad\quad\,
  Namhyuk Ahn$^1$ \qquad\quad\,
  Jangho Kim$^{2\dagger}$
  \\
  $^1$NAVER WEBTOON AI \\
  $^2$School of Artificial Intelligence, College of Computer Science, Kookmin University \\
}

\begin{document}

\maketitle

\begin{abstract}
Existing pruning methods utilize the importance of each weight based on specified criteria only when searching for a sparse structure but do not utilize it during training. In this work, we propose a novel approach - \textbf{M}agnitude \textbf{A}ttention-based Dynamic \textbf{P}runing (MAP) method, which applies the importance of weights throughout both the forward and backward paths to explore sparse model structures dynamically. Magnitude attention is defined based on the magnitude of weights as continuous real-valued numbers enabling a seamless transition from a redundant to an effective sparse network by promoting efficient exploration. Additionally, the attention mechanism ensures more effective updates for important layers within the sparse network. In later stages of training, our approach shifts from exploration to exploitation, exclusively updating the sparse model composed of crucial weights based on the explored structure, resulting in pruned models that not only achieve performance comparable to dense models but also outperform previous pruning methods on CIFAR-10/100 and ImageNet.

\end{abstract}

\section{Introduction}

Deep neural networks (DNNs) have demonstrated exceptional performance across various machine learning domains.
However, DNNs are often over-parameterized, resulting in high memory and computational costs during inference~\cite{zhang2021understanding}. To address this issue, model pruning has been extensively studied, aiming to remove redundant weights or neurons and create sparse models with reduced memory and floating-point operations (FLOPs) compared to dense ones~\cite{zhang2018learning,lecun1989optimal,singh2020woodfisher,chen2022state}.

Existing pruning methods typically determine the importance of weights (or filters) based on specific criteria and impose sparsity either before, after, or during training. Then, they select a sparse model based on weight importance~\cite{han2015learning}. Recently, dynamic model pruning methods have emerged, which dynamically update both important and unimportant weights to further explore the sparse structure~\cite{Lin2020Dynamic,guo2016dynamic}. However, these methods update all weights with the same magnitude throughout the training process, resulting in performance degradation due to the update of unimportant weights. Moreover, they do not differentiate the importance of weights in the forward and backward propagation during the learning process. Unlike conventional dense model training, where weight importance is not considered, pruning methods need to incorporate weight importance into their process because all pruning methods select the sparse structure based on their weight importance defined by criteria.

In this work, we propose a novel approach for model pruning, dubbed as \textbf{M}agnitude \textbf{A}ttention-based dynamic \textbf{P}running (MAP), which leverages the magnitude of model weights to identify more important connections in the dense network.
Our proposed MAP follows a gradual pruning schedule~\cite{zhu2017prune} so the method gradually increases the pruning ratio based on the current iteration (or epoch) until reaching the target pruning ratio (please see Section~\ref{Gradual}). By considering the on-the-fly element-wise importance of weights in the model, our magnitude attention provides a simple yet effective criterion for identifying important connections without additional costs.

We define the magnitude attention based on the current pruning ratio and the magnitudes of weights. This attention is implemented during both forward and backward paths of the model. In the forward pass, attention is given to important layers, guiding the model's update in the direction led by these layers. During the backward pass, even non-critical weights are dynamically updated to facilitate a transition to a sparse structure, while also giving more attention to crucial weights for more substantial updates. 
Furthermore, to promote the exploration and exploitation effect of the sparse model, we introduce two distinct phases.
In the \textit{exploration phase}, we apply magnitude attention to the search for the sparse structure, while in the \textit{exploitation phase}, we focus on improving the performance and robustness of the discovered sparse structure.

Our algorithm is designed to achieve two main objectives: \textbf{(1)} incorporating magnitude attention for model structure exploration, and \textbf{(2)} enabling exploitation of the explored sparse structure.

\begin{figure}[t]
\centering
 \includegraphics[width=1.0\linewidth]{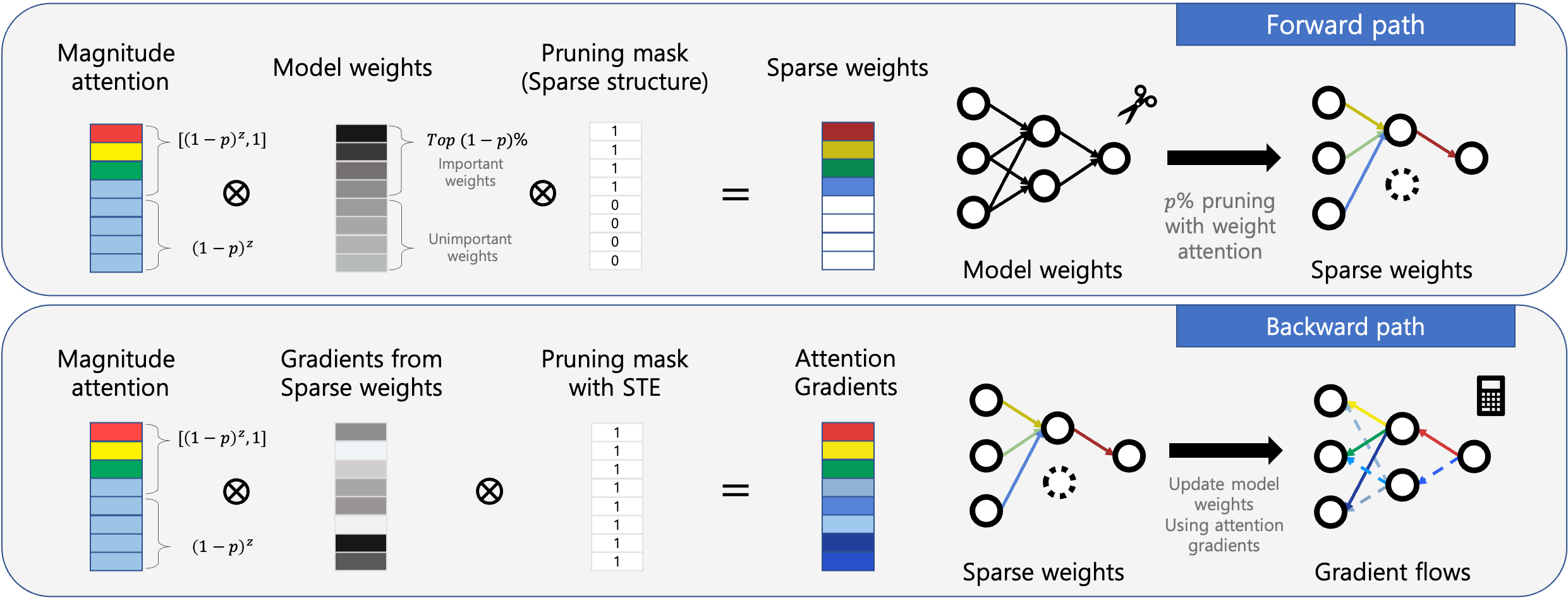}
  \caption{\textbf{Magnitude attention mechanism.} 
For clear understanding, the weights of the model are sorted in descending order based on their magnitudes. Unlike conventional pruning methods, MAP multiplies continuous values based on the magnitude of weights in the forward and backward paths. Magnitude attention assigns $(1-p)^z$ to unimportant (pruned) weights, while it assigns values ranging from $(1-p)^z$ to 1 based on the magnitude of important weights, where $p$ is a pruning ratio and $z$ is an attention strength value. During the backward path, the values of the mask corresponding to unimportant weights are set to 1 in order to facilitate exploration via straight-through estimator~\cite{bengio2013estimating}.}
  \label{fig:overall_framework}
\end{figure}

\textbf{Exploration via Magnitude-Based Attention.\;\;} 
During the forward and backward paths, weights are calculated based on their importance, with significant weights receiving more attention. Conversely, less important weights receive smaller updates during the backward path.
In the early training stages, both important and unimportant weights are updated similarly to facilitate dynamic structure searching (exploration). As the pruning rate increases, the focus shifts to emphasizing updates on important weights while reducing attention to unimportant ones. Our attention mechanism, which enables continuous and differential updates across layers, encourages effective exploration and ensures thorough consideration of each layer (Figure~\ref{fig:overall_framework}).

\textbf{Exploitation with Explored Sparse Structure.\;\;\;}
In the later stages of training, we shift focus from exploration to exploitation. This involves discontinuing the search for potential sparse networks and emphasizing the optimization of the sparse model by maintaining the well-established structure of the discovered sparse networks. Furthermore, we update the model by considering the directions influenced by the layers with the most significant impact on performance (Figure~\ref{fig:phase}).

The contribution of our work is summarized as follows:
\vspace{-0.9em}
\begin{itemize}[leftmargin=*]
    \item We propose a novel magnitude attention-based pruning method (Section~\ref{sec:ours}) and demonstrate its effectiveness, comparing it with competitive state-of-the-art pruning methods (Section~\ref{ex_results}).

    \item We analyze the impact of our magnitude attention and provide insights on the importance of choosing the appropriate mask or weighting scheme in the dynamic pruning (Section~\ref{learning}).
    
    \item We introduce an \textit{exploration-exploitation} strategy on dynamic pruning (Section~\ref{exploitation}). Notably, we observe that this yields highly effective results compared to other static\footnote{We refer to the approach of updating only the important weights as \textit{static} pruning, which contrasts with the concept of \textit{dynamic} pruning~\cite{Lin2020Dynamic,guo2016dynamic}.} pruning methods.
    
\end{itemize}

\section{Related Work}

\textbf{Sparsify-after-training} is the common strategy in the model pruning field. Firstly, it trains a dense model with $T$ epochs for convergence and then prunes the dense model with a target sparsity. To compensate for the accuracy drops, the sparse model is re-trained with a few epochs. Recent studies~\cite{renda2020comparing,le2021network,zimmer2023how} have highlighted the importance of the learning rate schedule in the retraining phase and have proposed specific guidelines for selecting an appropriate schedule for iterative magnitude pruning (IMP)~\cite{han2015learning} and provide valuable insights into the impact of learning rate schedules. 

\textbf{Sparsify-before-training} searches the sparse model at an early phase of the training and trains the sparse model with fewer FLOPS compared to a dense model~\cite{hayou2020robust,alizadeh2022prospect,lee2018snip}. Another line of research, Lottery Ticket Hypothesis (LTH)~\cite{frankle2018lottery} has demonstrated that using an appropriate initialization based on the sparse structure found through pruning algorithms is comparable to maintaining the performance achieved in the over-parameterized state.

\textbf{Sparsify-during-training} simultaneously searches and trains the sparse model from scratch in the training phase so this does not need a re-training phase. 
Many of these works rely on the reparameterization of weights through the utilization of either a binary mask or an element-wise nonlinear mapping~\cite{chen2023unified,zhou2021effective,liu2021sparse,NEURIPS2019_f34185c4,8578988,kim2020position}. Also, some studies search for a better sparse structure in the whole training where dead (pruned) weights can be recovered and an opposite case can be possible for making a better sparse structure. \textit{Dynamic model pruning}~\cite{Lin2020Dynamic,guo2016dynamic} is a type of sparsify-during-training approach. They update even the unimportant weights that do not participate in the forward path.
To do that, they utilize a straight-through estimator (STE)~\cite{bengio2013estimating} for gradient approximation. Through this method, it becomes possible to explore a more dynamic sparse structure. 

Our proposed MAP employs STE, similar to dynamic model pruning, to update unimportant weights. However, in contrast to the previous dynamic pruning method, we introduce magnitude attention and an exploration-exploitation strategy to selectively train weights and boost the learning stability.

\section{Backgrounds}
\subsection{Dynamic Model Pruning}
The pruning process eliminates unimportant weights based on a specified criterion, making a sparse model. This is exemplified by magnitude-based pruning, which sorts the weights and assigns a zero mask to the smallest magnitude weights until the desired pruning ratio is met. Upon calculating the pruning mask, $M = [m_1, ... m_n] \subset \mathbb{R}^n$ where $M \in \{0,1\}^n$, we can build the sparse model $\overline W= M \odot W$ by multiplying the pruning mask with the model weights $W = [w_1, ... w_n] \subset \mathbb{R}^n$. Taking into account the pruning ratio $P$, the given dataset $D$ with individual samples $(x_k,y_k)$, and the sparse model parameterized by $M\odot W$, the optimization problem can be formulated as follows:
\begin{equation}
\min\limits_{W}\frac{1}{B}\sum\limits_{i}^{B}\mathcal{L}(f({M\odot W;x_k}),y_k) \quad s.t. \quad \|W\|_{0} \leq (1-P)\cdot n
\label{eq:objective}
\end{equation}
where, $B$ is batch size and $n$ is \# weights. $\|\cdot\|_{0}$ denotes $L_0$ norm. Dynamic pruning~\cite{Lin2020Dynamic,guo2016dynamic} not only trains the sparse model but also jointly evolves the unimportant weights with the straight-through estimator (STE)~\cite{bengio2013estimating}. The update rule of $j$-th weight with objective function Eq.~(\ref{eq:objective}) is as follows:
\begin{equation}
    {w^{i+1}_j} = {w^i_j} - \eta 
      \frac{\partial \mathcal{L}}{\partial \overline{w}^i_j}\frac{\partial \overline{w}^i_j}{\partial w^i_j}  \approx {w^i_j} - \eta 
      \frac{\partial \mathcal{L}}{\partial \overline{w}^i_j},  \:\: \forall j \in \{1,\dots,n\} 
      %\forall w_i \in \mathbf{W} 
\label{Dynamic_STE}
\end{equation}
where $i$ is an iteration. When the pruning mask of $j$-th weight is zero i.e. ${\partial \overline{w}^i_j}/{\partial w^i_j}=m_j=0$, then, unimportant weight ($w^i_j$) cannot be updated. However, dynamic model pruning methods approximate  ${\partial \overline{w}^i_j}/{\partial w^i_j}=1$ to update unimportant weights for searching the structure of sparse model dynamically.

\subsection{Gradual Pruning}
\label{Gradual}
In contrast to one-shot pruning, which prunes the model to achieve a target pruning ratio in a single shot and can result in performance degradation without retraining, gradual pruning increases the current pruning ratio incrementally based on the current epoch~\cite{zhu2017prune}. Gradual pruning is widely employed in sparsity-during-training since they better discover a dynamic sparse structure than a one-shot manner. In this approach, the current pruning ratio ($P_c$) is gradually increased from an initial pruning ratio ($P_s$ at $c=c_0$) to a target pruning ratio ($P_t$) as the current epoch ($c$) progresses.
\begin{equation} 
P_c = P_t + (P_s-P_t)(1-\frac{c-c_0}{E})^3, \quad c \in \{c_0, ..., c_0+E\}.
\label{eq:sparsity_ratio}
\end{equation} 
where $E$ is the total epochs. We adopt this gradual pruning schedule into our method.

\section{Magnitude Attention-based Dynamic Model Pruning}
\label{sec:ours}
As we adopt both dynamic and gradual model pruning, the pruning ratio changes in each epoch. Following Lin \textit{et al.}~\cite{Lin2020Dynamic}, we also modify the mask $M = [m_1, ... m_n] \subset \mathbb{R}^n$ for each $\mathcal{F}$ iteration within the same epoch to search the sparse structure. Consequently, we can define the pruning mask based on the current iteration ($i$) and epoch ($c$).
We note that the pruning ratio depends on the current epoch and the pruning mask is updated every $F$ iteration based on the threshold at $i$-th iteration. The pruning threshold $\lambda^{(i,{P_c})}$ is computed according to the current pruning rate based on the epoch (Eq.~(\ref{eq:sparsity_ratio})). The pruning mask of $j$-th weight at the $i$-th iteration, $m^{(i,P_c)}_{j}$, can be calculated as follows:
\begin{equation}
\label{calc:mask}
\begin{split}
    &m^{(i,P_c)}_{j} = \begin{cases}
    Mask(\lambda^{(i,{P_c})},w^i_j) & \text{if } \quad i \;\%\; \mathcal{F}==0 \\
     m^{(i-1,P_c)}_{j} & \text{elsewise}, \end{cases}\\
  &Mask(\lambda^{(i,{P_c})},w^i_j) = \begin{cases}
    1 & \text{if } \quad \lvert w^i_j \rvert> \lambda^{(i,{P_c})} \\
     0 & \text{elsewise}, 
    \end{cases}
\end{split}
\end{equation}
The proposed MAP consists of a magnitude attention-based exploration and exploitation upon an explored sparse structure.
We explain two parts in the following sections.

\subsection{Exploration with a Magnitude Attention}
In this work, we develop our magnitude attention approach based on three principles. \textbf{(1)} During the early stages of learning, due to a low pruning ratio, we facilitate effective exploration by updating both important and unimportant weights with similar magnitudes. As the pruning ratio increases, we shift the focus towards updating important weights rather than unimportant ones. \textbf{(2)} We distinguish and update important weights based on their magnitudes in the forward and backward paths. \textbf{(3)} To further enhance the exploration, we treat the attention values of unimportant weights uniformly and assign these to be contiguous with the ones of important weights.

To adhere to these three principles, in forward path, we multiply important weights by normalized values, ranging in $[(1-P_c)^z, 1]$, while unimportant weights are removed by a zero-valued mask.
In backward path, at the beginning of training when the pruning ratio is low, both important and unimportant weights are updated similarly.
As learning progresses, we concentrate on the identified (sparse) structure by decreasing the update degree of unimportant weights, which helps to prioritize and refine the most relevant connections, leading to improved performance. 
At every $i$-th iteration, the magnitude attention value $A^i = [a^{i}_1, ..., a^{i}_n]$ is applied to the model weights. The value $a^{i}_{j}$, which corresponds to the $j$-th weight ${w^{i}_j}$, is defined as follows:
\begin{equation}
\label{attention}
\begin{split}
    &a^{i}_{j} = \begin{cases}
    Att(w^{i}_{j},m^{(i,P_c)}_{j}) & \text{if } \quad i \;\%\; \mathcal{F}==0 \\
     a^{i-1}_{j} & \text{elsewise}, 
    \end{cases} \\
    &Att(w^{i}_{j},m^{(i,P_c)}_{j}) = \begin{cases}
    \frac{\lvert w^i_j \rvert-min(\lvert W^i \rvert)}{max(\lvert W^i \rvert) - min(\lvert W^i \rvert)}\cdot(1-(1-P_c)^z)+(1-P_c)^z & \text{if } \quad m^{(i,P_c)}_{j}  ==1 \\
     (1-P_c)^z & \text{elsewise}, 
    \end{cases}
    \end{split}
\end{equation}
where $max(\lvert W^i \rvert)$ and $min(\lvert W^i \rvert)$ are the element-wise absolute max and min values of the model weights at $i$-th iteration ($W^i$). $z$ is an attention strength value that controls the range of attention. We will discuss the effect of $z$ in Suppl. As depicted in Figure~\ref{fig:overall_framework}, magnitude attention of important weights are in the range of $[(1-P_c)^z\;, 1]$ and unimportant weights are calculated by $(1-P_c)^z$. 

In forward propagation, we multiply magnitude attention to the sparse model (i.e. $\overline{w}^{i}_j=m^{(i,P_c)}_{j}\;\cdot\;{w}^{i}_j$ and $\widetilde{w}^{i}_j=a^{i}_{j}\;\cdot\;\overline{w}^{i}_j$) and calculate the loss function to update model weight. As a result, the update rule of the proposed magnitude attention-based dynamic model pruning can be written as follows:
\begin{equation}
    {w^{i+1}_j} = {w^{i}_j} - \eta 
      \frac{\partial \mathcal{L}}{\partial \widetilde{w}^{i}_j}\frac{\partial \widetilde{w}^{i}_j}{\partial \overline{w}^{i}_j}\frac{\partial \overline{w}^{i}_j}{\partial {w^{i}_j}}  \approx {w^{i}_j} - \eta 
      \frac{\partial \mathcal{L}}{\partial \widetilde{w}^{i}_j}\cdot a^i_j,  \:\: \forall j \in \{1,\dots,n\} 
      %\forall w_i \in \mathbf{W} 
\label{MAP_calculation}
\end{equation}
In backward propagation, as depicted in Eq.~(\ref{MAP_calculation}), the attention value is multiplied by gradients. Figure~\ref{fig:overall_framework} shows an overview of the exploration phase with magnitude attention considering the importance of weights in forward and backward paths. 
This double-sided attention mechanism places a higher priority on updating layers that have a substantial effect on the model's performance.
By concentrating on the critical layers and providing them with targeted updates, we can enhance the overall performance of the model. For a detailed analysis, please refer to the Suppl.

\subsection{Exploitation with an Explored Sparse Structure}
\label{exploitation}

\begin{wrapfigure}{r}{65mm}
\vspace{-1.5em}
\includegraphics[width=1\linewidth]{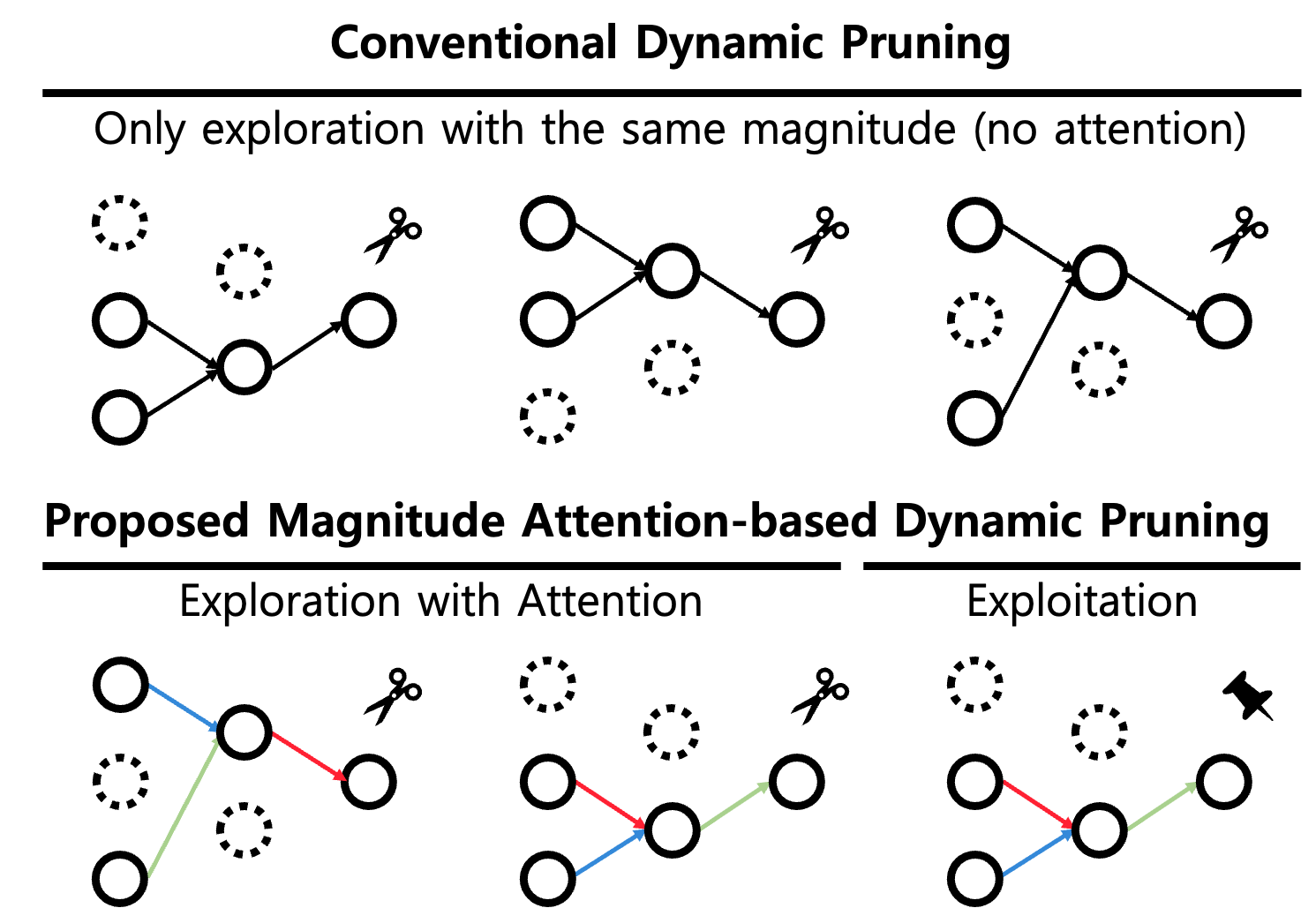}
  \caption{\textbf{MAP vs. conventional dynamic pruning} in a training timeline perspective.}
  \label{fig:phase}
  \vspace{-1em}
\end{wrapfigure}

Typical dynamic pruning methods employ STE~\cite{bengio2013estimating} to uncover sparse structures throughout the entire training process. While this approach is more effective than methods that disregard the updating of unimportant weights, it suffers learning instability due to the gradient approximation effect of STE. We found that fixing the sparse structure and updating only the values of important weights is highly effective for dynamic model pruning methods with dynamic search.
We conjecture that such strategy elaborates further \textit{exploitation} and eventually, well-explored sparse models can receive improvements on both performance and robustness (Section~\ref{analysis_exploitation}).
Figure~\ref{fig:phase} compares conventional dynamic model pruning to ours in terms of the entire training timeline.
MAP first explores the sparse model with a novel magnitude attention module, then after $K$ epochs, this ceases updating attention values and sparse masks to solely train fixed sparse models.
Algorithm~\ref{algo:map} outlines the overall process of the proposed MAP method.

\section{Experiments}

\begin{algorithm}[t]
\centering
\caption{\text{Magnitude Attention-based Dynamic Pruning (MAP)}}
\label{algo:map}
\begin{algorithmic}[1]
\label{algorithm}
\Require Mask update frequency: $\mathcal{F}$, Pruning mask: ${M}\in\{0, 1\}^n$, Model weights: ${W} \in \mathbb{R}^n$, Total iteration: $T$. Target exploitation iteration: $K$ \Comment{$n$ is the number of parameters}
\For{$i$ = 1 ,..., $T$}
\Comment{$i$ is the index of iteration}
\If {$i$ $\%$ $\mathcal{F}==0$ \: \textbf{and} \: $i$ $\leqslant$ $K$} \Comment{After $K$ iteration, it enters the exploitation phase.}
\State Update pruning mask ${M}^i$ with Eq.(\ref{calc:mask}) 
\State Update magnitude attention ${A}^i$ with Eq.(\ref{attention})
\EndIf
\State Compute a sparse model $\overline{{W}}^i={M}^i \odot W^i$
\Comment{$\odot$ represents the hardmard product}
\State Apply magnitude attention to the sparse model $\widetilde{{W}}^i=A^i \odot \overline{{W}}^i$
\State Update weights ${W^{i+1}}  = {W^{i}} - \eta\frac{\partial \mathcal{L}}{\partial \widetilde{{W}}^i}\odot A^i$ 
\EndFor
\end{algorithmic}
\end{algorithm}

\begin{table}[t]
\centering
\caption{\textbf{Comparison of pruning methods.} Here, we report Top-1 test accuracy by selecting the highest score achieved for each method. Underlined values indicate instances where the pruned model outperforms the dense one. N/A denotes the method cannot converge.}
\small
\begin{adjustbox}{valign=t}
\begin{tabular}{l|cccc}
\toprule
\textbf{Backbone} & \multicolumn{4}{c}{\textbf{ResNet-20}} \\
\midrule
Sparsity & 70\% & 80\% & 90\% & 95\% \\
\midrule\midrule
\textbf{Dense} & \multicolumn{4}{c}{\perf{92.36}{0.10}} \\
SM~\cite{dettmers2019sparse} & \perf{91.98}{0.01} & \perf{91.54}{0.16} & \perf{89.76}{0.40} & \perf{83.03}{0.74} \\
DSR~\cite{mostafa2019parameter} & \perf{92.00}{0.19} & \perf{91.78}{0.28} & \perf{87.88}{0.04} & N/A \\
DPF~\cite{Lin2020Dynamic} & \perf{92.42}{0.14} & \perf{92.17}{0.21} & \perf{90.88}{0.07} & \perf{88.01}{0.30} \\
CP~\cite{srinivas2022cyclical} & - & - & \perf{90.90}{0.10} & \perf{89.29}{0.20} \\
MAP (ours) & \perf{\textbf{\underline{92.72}}}{0.20} & \perf{\textbf{92.30}}{0.13} & \perf{\textbf{92.19}}{0.04} & \perf{\textbf{90.64}}{0.32} \vspace{0.4em}\\
\toprule
\textbf{Backbone} & \multicolumn{2}{c}{\textbf{ResNet-32}} & \multicolumn{2}{c}{\textbf{ResNet-56}} \\
\midrule
Sparsity & 70\% & 80\% & 70\% & 80\% \\
\midrule\midrule
\textbf{Dense} & \multicolumn{2}{c}{\perf{93.22}{0.07}}  & \multicolumn{2}{c}{\perf{94.34}{0.19}} \\
MAP (ours) & \perf{\underline{93.32}}{0.17} & \perf{\underline{93.26}}{0.23} & \perf{\underline{94.35}}{0.04} & \perf{93.90}{0.16} \\
\midrule
Sparsity & 90\% & 95\% & 90\% & 95\% \\
\midrule\midrule
\textbf{Dense} & \multicolumn{2}{c}{\perf{93.22}{0.07}}  & \multicolumn{2}{c}{\perf{94.34}{0.19}} \\
SNIP~\cite{lee2018snip} & \perf{90.40}{0.26} & \perf{87.23}{0.29} & - & - \\
SM~\cite{dettmers2019sparse} & \perf{91.54}{0.18} & \perf{88.68}{0.22} & \perf{92.73}{0.21} & \perf{90.96}{0.40} \\
DSR~\cite{mostafa2019parameter} & \perf{91.41}{0.23} & \perf{84.12}{0.32} & \perf{93.78}{0.20} & \perf{92.57}{0.09} \\
DPF~\cite{Lin2020Dynamic} & \perf{92.42}{0.18} & \perf{90.94}{0.35} & \perf{93.95}{0.11} & \perf{92.74}{0.08} \\
GMP~\cite{gupta2022complexity} & \perf{92.67}{0.03} & \perf{90.65}{0.13} & - & - \\
BIMP~\cite{zimmer2023how} & - & - & \perf{93.35}{0.13} & \perf{92.57}{0.32} \\
MAP (ours) & \perf{\textbf{93.13}}{0.03} & \perf{\textbf{92.16}}{0.11} & \perf{\textbf{94.02}}{0.04} & \perf{\textbf{93.17}}{0.13} \\
\bottomrule
\end{tabular}
\end{adjustbox}
\hfill
\begin{adjustbox}{valign=t}
\begin{tabular}{l|cc}
\toprule
\textbf{Backbone} & \multicolumn{2}{c}{\textbf{ResNet-50}} \\
\midrule
Sparsity & 80\% & 90\% \\
\midrule\midrule
\textbf{Dense} & \multicolumn{2}{c}{75.95} \\
Incre.~\cite{zhu2017prune} & 74.25 & 73.36 \\
SM~\cite{dettmers2019sparse} & 74.59 & 72.65  \\
DSR~\cite{mostafa2019parameter} & 73.30 & 71.60 \\
STR~\cite{kusupati2020soft} & 70.70 & 70.13 \\
GSM~\cite{NEURIPS2019_f34185c4} & 72.75 & 70.08 \\
LC~\cite{8578988} & 73.87 & 67.57 \\
DST~\cite{liu2020dynamic} & 73.16 & 71.35 \\
GMP~\cite{gupta2022complexity} & 74.19 & 72.80 \\
BIMP~\cite{zimmer2023how} & 75.08 & 73.53 \\
DPF~\cite{Lin2020Dynamic} & 75.12 & 74.38 \\
DNW~\cite{wortsman2019discovering} & 75.27 & 74.29 \\
CP~\cite{srinivas2022cyclical} & 75.30 & 73.30 \\
MAP (ours) & \textbf{75.90} & \textbf{74.90} \\
\bottomrule
\end{tabular}
\end{adjustbox}
\label{table:comparison}
% \vspace{-1.5em}
\end{table}

\subsection{Experimental setup}
\noindent\textbf{Datasets \& Backbones.} We evaluate the performance of pruning methods on three datasets, CIFAR-10/100~\cite{krizhevsky2009learning} and ImageNet~\cite{deng2009imagenet}. All the methods prune ResNet~\cite{he2016deep}, specifically utilizing 20, 32, and 56-layers of ResNet for CIFAR-10/100 while employing 50 layers version for ImageNet assessments.
For the CIFAR-10/100 benchmarks, which include comparison and model analysis, we present the average accuracy obtained from three runs.

\noindent\textbf{Baselines.} Our model is compared against various state-of-the-art pruning methods including SM~\cite{dettmers2019sparse}, SNIP~\cite{lee2018snip}, DSR~\cite{mostafa2019parameter}, DPF~\cite{Lin2020Dynamic}, CP~\cite{srinivas2022cyclical}, GMP~\cite{gupta2022complexity}, BIMP~\cite{zimmer2023how}, SET~\cite{Constantin2018scalable}, Deep-R~\cite{DBLP:conf/iclr/BellecK0L18}, GraSP~\cite{Wang2020Picking}, STR~\cite{kusupati2020soft}, GSM~\cite{NEURIPS2019_f34185c4}, LC~\cite{8578988}, DST~\cite{liu2020dynamic}, and DNW~\cite{wortsman2019discovering}.

\noindent\textbf{Implementation details.} To conduct a rigorous comparison, we meticulously followed the experimental setup specified in DPF~\cite{Lin2020Dynamic} for CIFAR-10/100 and STR~\cite{kusupati2020soft} for ImageNet.
Our implementation incorporates magnitude-based unstructured weight pruning~\cite{han2015learning} throughout all layers of the neural network, consciously avoiding layer-specific pruning. We apply the proposed pruning method and others to the entire network except for the batch normalization~\cite{ioffe2015batch} and final fully connected layers.
In Suppl. we describe more details on our methods including hyperparameters and training setups.

\begin{wraptable}{r}{0.35\linewidth}
\caption{\textbf{Methods comparison} on CIFAR-100 (Top-1 Accuracy).}
\vspace{-0.6em}
\small
\centering
\begin{tabular}{l|cc}
\toprule
\textbf{Backbone} & \multicolumn{2}{c}{\textbf{ResNet-32}} \\
\midrule
Sparsity & 90\% & 95\% \\
\midrule\midrule
\textbf{Dense} & \multicolumn{2}{c}{74.64} \\
DSR~\cite{mostafa2019parameter} & 69.63 & 68.20 \\
SET~\cite{Constantin2018scalable} & 69.66 & 67.41 \\
Deep-R~\cite{DBLP:conf/iclr/BellecK0L18} & 66,78 & 63.90 \\
SNIP~\cite{lee2018snip} & 68.89 & 65.22 \\
GraSP~\cite{Wang2020Picking} & 69.24 & 66.50 \\
\midrule
\textbf{Dense} & \multicolumn{2}{c}{70.60} \\
DPF~\cite{Lin2020Dynamic} & 67.15 & 62.97 \\
MAP (ours) & \textbf{70.42} & \textbf{67.23} \\
\bottomrule
\end{tabular}
\label{table:cifar100comparison}
\end{wraptable}

\subsection{Quantitative Comparison}
\label{ex_results}
Table~\ref{table:comparison} (left) presents the top-1 test accuracy on CIFAR-10. Our proposed method, MAP, consistently outperforms its competitors, demonstrating the effectiveness of magnitude attention. For instance, compared to the second-best methods (e.g., CP~\cite{srinivas2022cyclical} and DPF~\cite{Lin2020Dynamic}) at 95\% sparsity, MAP shows a performance improvement of over 1\% for both the 20 and 32-layer ResNet and a 0.4\% increase in ResNet-56. Notably, at low sparsity levels (e.g., 70\% or 80\%), MAP surpasses the performance of dense networks.
As depicted in Table~\ref{table:cifar100comparison}, MAP exceeds the other pruning methods on CIFAR-100 as well, similar to the CIFAR-10 benchmark.
MAP also outperforms dense network in low sparsity cases in CIFAR-100; our method achieves 70.93\% in 80\% sparsity while dense network shows 70.60\%.
We provide additional experimental results in Suppl.

In Table~\ref{table:comparison} (right) we compare various pruning methods in the ImageNet dataset at sparsity levels of 80\% and 90\%. Our proposed MAP consistently delivers outstanding accuracy, outperforming other methods by approximately 0.52\% points (e.g., 74.38 vs. 74.90) in the 90\% sparsity case. Furthermore, even at 80\% pruning, MAP's performance is nearly on par with the dense baseline (75.95 vs. 75.90), underscoring the effectiveness of our proposed MAP method.

\subsection{Model Analysis}
In this section, we delve into the analysis of our proposed model, focusing on two key aspects that contribute to its overall performance.
We begin by exploring the learning mechanism of MAP in Section~\ref{learning}, where we dissect the importance of mask in forward path and weight term in backward path to make an effective pruning method.
In Section~\ref{analysis_exploitation}, we examine the efficacy of the exploitation phase within the dynamic pruning approach.
Unless mentioned, we conduct analysis using the ResNet-56 backbone on the CIFAR-100 dataset with a pruning ratio of 90\%. In Suppl., we include additional results on diverse training setups with different backbones and datasets.

\begin{table}[t]
    \caption{\textbf{Pruning configuration in the internal study.} We manipulate a weight mask in the forward path and employ different weighting schemes for sparse and non-sparse models in the backward path.}
    \centering
    \begin{tabular}{c|ll}
    \toprule
    Config & Forward path & Backward path \\
    \midrule\midrule
    A & Sparse with a binary mask (Eq.\ref{calc:mask}). & Sparse with a binary mask (Eq.\ref{calc:mask}). \\
    \midrule
    B & Sparse with a binary mask (Eq.\ref{calc:mask}). & All layers (sparse and non-sparse) equally. \\
    \midrule
    C & Sparse with a binary mask (Eq.\ref{calc:mask}). & Sparse with full weight, non-sparse with  $(1-p_c)^z$. \\
    \midrule    
    D & Sparse with an attention mask (Eq.\ref{attention}). & Sparse with Eq.\ref{attention}, non-sparse with $(1-p_c)^z$. \\
    \bottomrule
    \end{tabular}
    \label{table:config}
\end{table}

\subsubsection{Understanding the Learning Mechanism of MAP}
\label{learning}

As depicted in Table~\ref{table:config}, we examine four distinct models to analyze the learning mechanism of MAP. For each setup, we differentiate between a mask for the sparse model, employed during the forward path, and a weighting scheme for both sparse and non-sparse models, utilized in the backward path.
\textsc{Config A} incorporates a binary mask for both forward and backward paths~\cite{zhu2017prune}.
\textsc{Config B} is the same as \textsc{Config A} in the forward path, but this utilizes both sparse and non-sparse in backward propagation. Note that DPF~\cite{Lin2020Dynamic} is the same mechanism as this setup.
\textsc{Config C} also uses a binary mask in the forward path, however, it employs $(1-p_c)^z$ weighting scheme for the non-sparse model in the backward path.
By doing so, it progressively decreases the concentration of the non-spare model when the pruning ratio is gradually increased.
\textsc{Config D} employs an attention mask (Eq.\ref{attention}) to the sparse model in the forward and backward path.
This elaborates the exploration effect, which is beneficial to find a new sparse model in the later part of the training.

\begin{figure}[h]
\begin{minipage}[t]{0.40\linewidth}
\centering
\vspace{-5.0em}
\begin{tabular}{ccc}
\toprule
\multirow{2}[2]{*}{Config} & \multicolumn{2}{c}{Accuracy} \\
\cmidrule{2-3}
& Last & Best \\
\midrule\midrule
Dense & \multicolumn{2}{c}{\perf{71.71}{0.64}} \\
A & \perf{68.84}{0.72} & \perf{69.33}{0.84} \\
B & \perf{65.32}{0.38} & \perf{70.44}{0.15} \\
C & \perf{69.77}{0.20} & \perf{70.60}{0.13} \\
D & \perf{65.92}{1.28} & \perf{70.93}{0.31} \\
\bottomrule
\end{tabular}
\end{minipage}
\quad
\begin{minipage}{0.56\linewidth}
\centering
\includegraphics[width=1\linewidth]{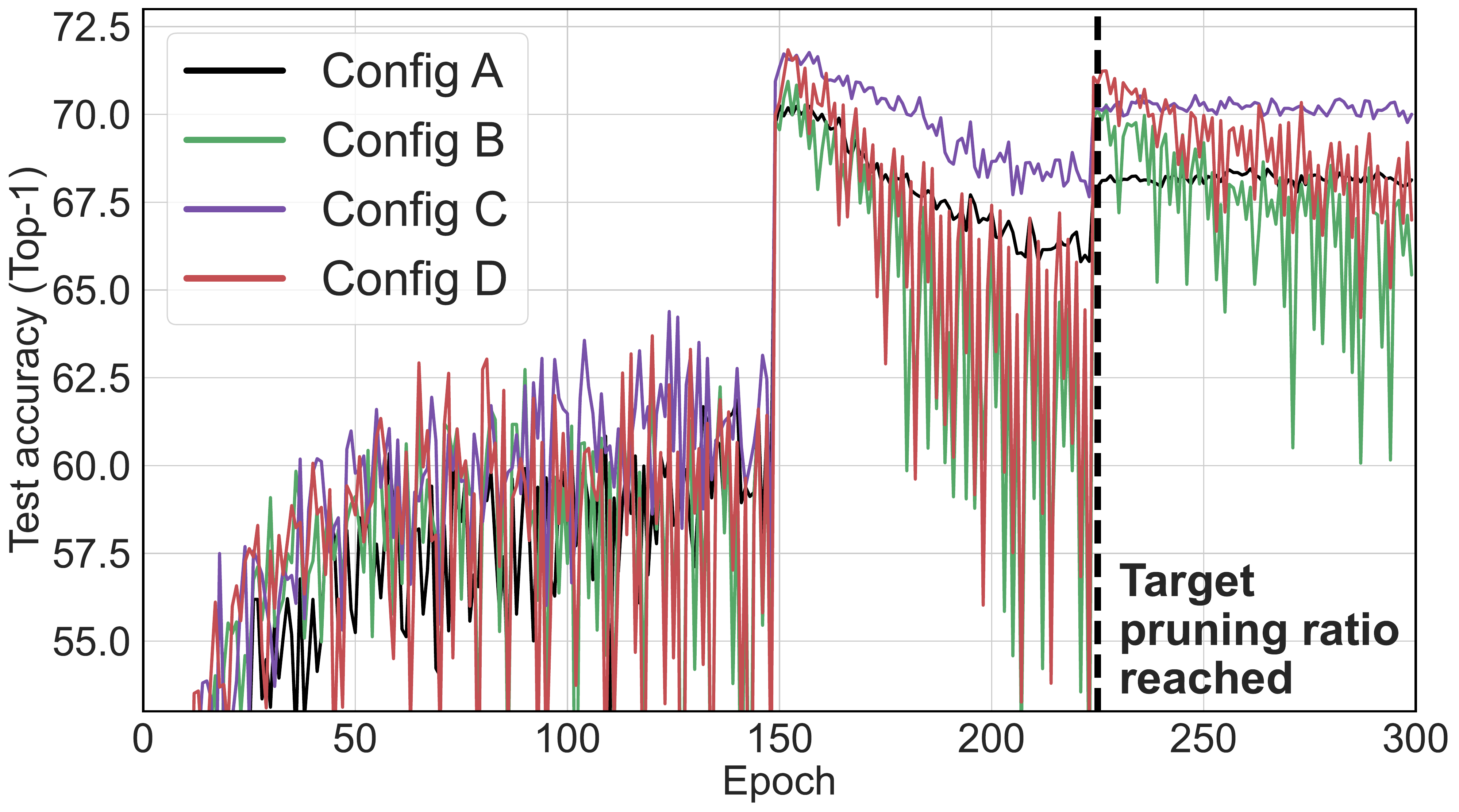}
\end{minipage}
\vspace{-0.5em}
\caption{\textbf{Training of \textsc{Config A-D}.} We train ResNet-56 on CIFAR-100. (\textbf{Left}) Top-1 test accuracy (best and last epochs) of each approach. (\textbf{Right}) Test accuracy during the training process.}
\label{fig:comp_abc}
\end{figure}

\paragraph{Weighting scheme for the non-sparse model.}
Although dynamic pruning (e.g. DPF~\cite{Lin2020Dynamic}), demonstrates its strengths in the exploration of sparse models, it often updates non-sparse models more frequently than necessary due to the equal weight update given to both sparse and non-sparse candidates.
To investigate this, we compare \textsc{Config B}, which has the same setting as DPF and \textsc{Config C}, which downweights the non-sparse model during the backward path using $(1-p_c)^z$ weighting scheme.
As shown in Figure~\ref{fig:comp_abc} (right), the test accuracy in the early stages of training ($<150$ epoch) are similar for both configurations as they explore the sparse candidates in the same manner.
However, a noticeable difference in the accuracy emerges in the mid-to-late stages of training ($>150$ epoch).

\begin{wrapfigure}{r}{0.45\linewidth}
\centering
\vspace{-1.2em}
\includegraphics[width=\linewidth]{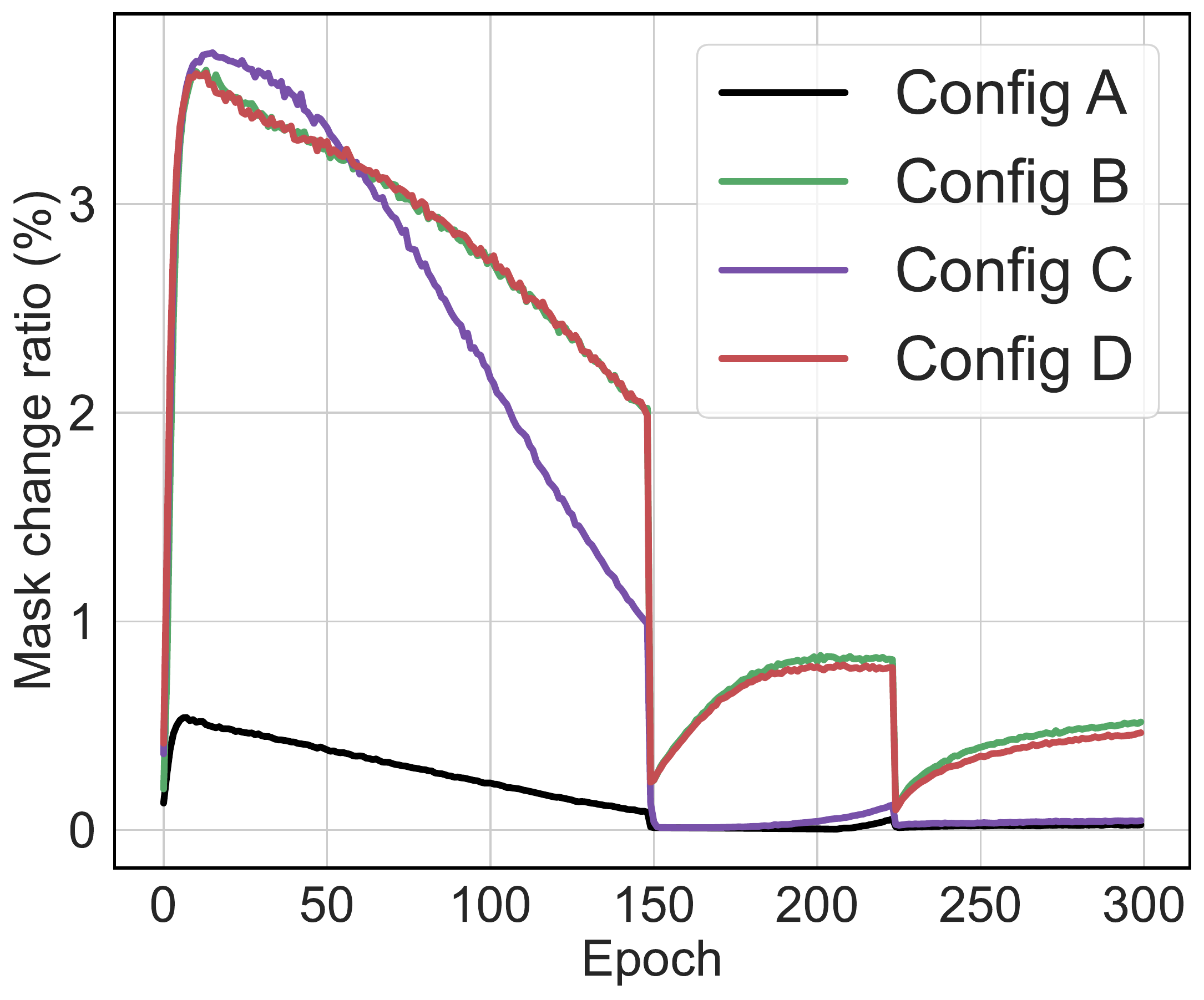}
\vspace{-1.5em}
\caption{\textbf{Ratio of pruning mask changes.} We visualize \textsc{Config A-D} (Section~\ref{learning})}
\label{fig:mask_abc}
\vspace{-2em}
\end{wrapfigure}

We conjecture that this occurs because we guide \textsc{Config C} to update sparse model more frequently at the later training phase, leading to stabilized learning.
Figure~\ref{fig:mask_abc} visualizes the extent of changes in the sparse candidates during the training process using the sparse mask change ratio.
While \textsc{Config B} consistently exhibits a high change ratio, \textsc{Config C} initially shows a high ratio that rapidly declines as training progresses.
Consequently, we can infer that \textsc{Config C} modifies the weighting scheme for non-sparse models in the backward path, thereby reducing the exploration effect.
Such alteration naturally strengthens the exploitation of sparse models (as we consider the exploration-exploitation trade-off), resulting in an overall improvement in model performance and robustness than \textsc{Config B} (and DPF~\cite{Lin2020Dynamic}).

\paragraph{Magnitude attention.}
\label{main/5.3.1.mag_att}
The weighting scheme for non-sparse models (\textsc{Config C}) exhibits higher performance and stability compared to \textsc{Config B}. Nevertheless, it has the drawback of significantly reducing the exploration effect in the later stages of training (Figure~\ref{fig:mask_abc}). Reducing the exploration power may limit the interchange between sparse and non-sparse models, which potentially poses the risk of getting stuck in local optima. In addition, our proposed exploitation method (Section~\ref{exploitation}) requires substantial exploration to achieve optimal results. To investigate an effective way of improving exploration without sacrificing performance, we compare \textsc{Config C} to \textsc{Config D}, which employs the proposed magnitude attention in both the forward and backward paths.

The magnitude attention allows a more active transition between candidates by making the update rule for sparse and non-sparse models a unified continuous form. Additionally, this approach sorts connections by importance based on the model weight's magnitude, allocating more attention to critical models. As shown in Figure~\ref{fig:mask_abc}, \textsc{Config D} successfully raises the low exploration of \textsc{Config C} to that of the original dynamic pruning method (\textsc{Config B}; DPF) while maintaining overall learning performance and stability (Figure~\ref{fig:comp_abc}). 

In essence, compared to the static or dynamic pruning baseline (\textsc{Config A} or B), our method (\textsc{Config D}), which incorporates magnitude attention and a weighting scheme for non-sparse models, achieves significant improvements in learning performance and stability while maintaining an equivalent exploration perspective. This advancement can be considered a substantial development, taking into account the exploration-exploitation trade-off.

\subsubsection{Understanding the Exploitation on Dynamic Pruning}
\label{analysis_exploitation}
In Section~\ref{exploitation}, we improved the exploitation effect on the dynamic pruning approach.
This method involves freezing the pruning mask in the later stages of training to prevent any changes in the sparse network, allowing only the sparse model to be updated while the final network structure remains fixed.
By maximizing exploitation in this manner, the overall performance and robustness of the learning process are significantly increased.

\begin{figure}[h]
\begin{minipage}[t]{0.40\linewidth}
\vspace{-3.9em}
\centering
\begin{tabular}{clc}
\toprule
\multirow{2}[2]{*}{Config} & \multicolumn{2}{c}{$\Delta$ Accuracy to w/o \textit{Exploit}} \\
\cmidrule{2-3}
& \;\;\; Last & Best \\
\midrule\midrule
A & + 0.02 & + 0.11 \\
B & + 6.26 & + 1.21 \\
C & + 0.40 & + 0.19 \\
D & + 6.10 & + 1.32 \\
\bottomrule
\end{tabular}
\end{minipage}
\quad
\begin{minipage}{0.56\linewidth}
\centering
\includegraphics[width=1\linewidth]{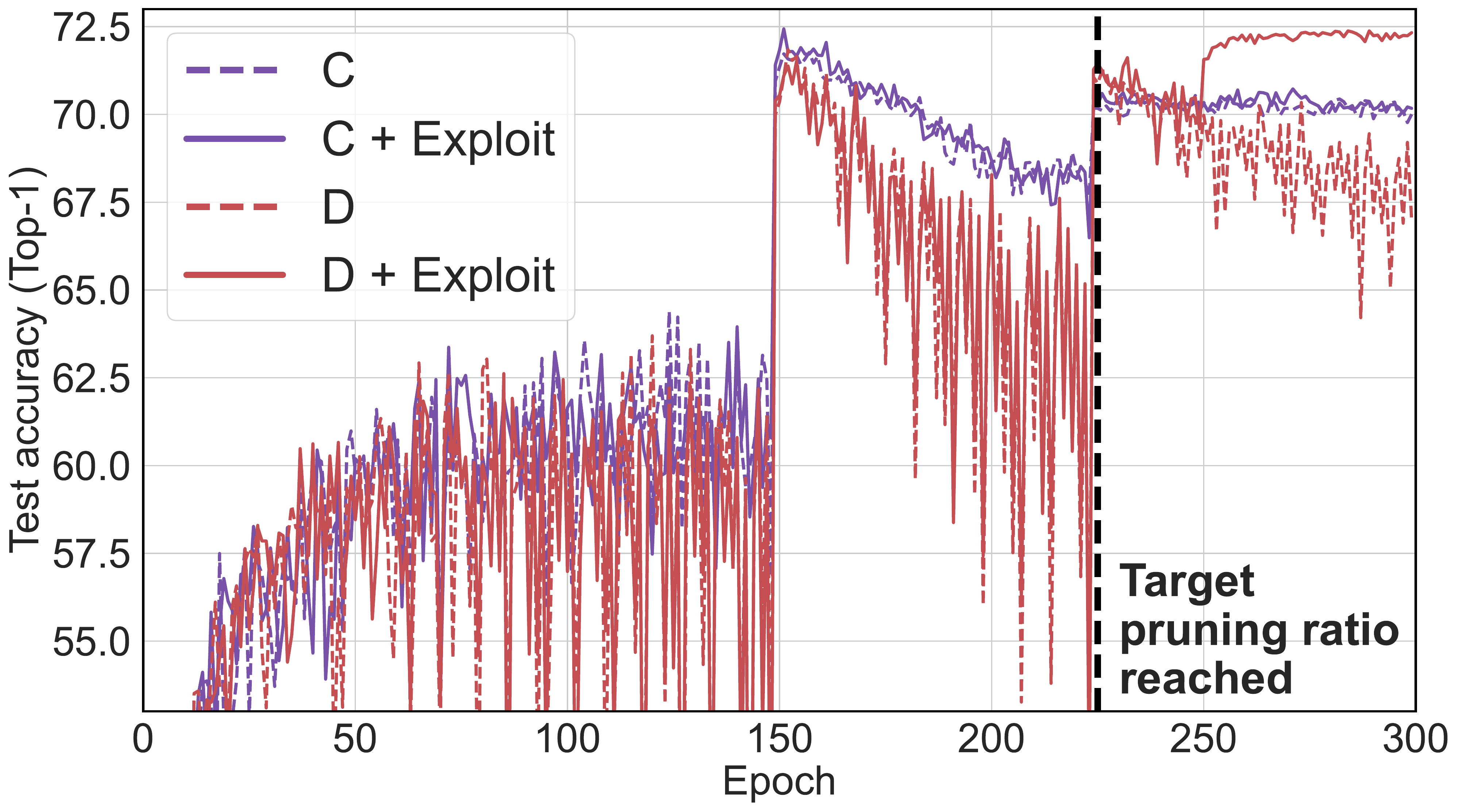}
\end{minipage}
\vspace{-0.5em}
\caption{\textbf{Training of \textsc{Config A-D} with exploitation.} (\textbf{Left}) The difference in accuracy between the w/ and w/o exploitation. (\textbf{Right}) Test accuracy during the training process of \textsc{Config C, D}.}
\label{fig:comp_bc_exploit}
\end{figure}

To demonstrate the efficacy of exploitation in dynamic pruning and how exploitation works with exploration, we conduct an experiment by applying the exploitation learning to multiple pruning setups (Figure~\ref{fig:comp_bc_exploit}). We note that by their design, \textsc{Config A, C} less explore and \textsc{Config B, D} enhances exploration. For example, \textsc{Config C} minimizes exploration to enhance exploitation while \textsc{Config D} elaborates both significantly.
Although further exploitation is introduced, \textit{less-explored} models (\textsc{Config A, C}) only show marginal performance improvement.
We argue that this is because their restricted exploration makes additional exploitation ineffective and even resulting overfitting.

\begin{wrapfigure}{r}{0.35\linewidth}
\centering
\vspace{1.3em}
\includegraphics[width=\linewidth]{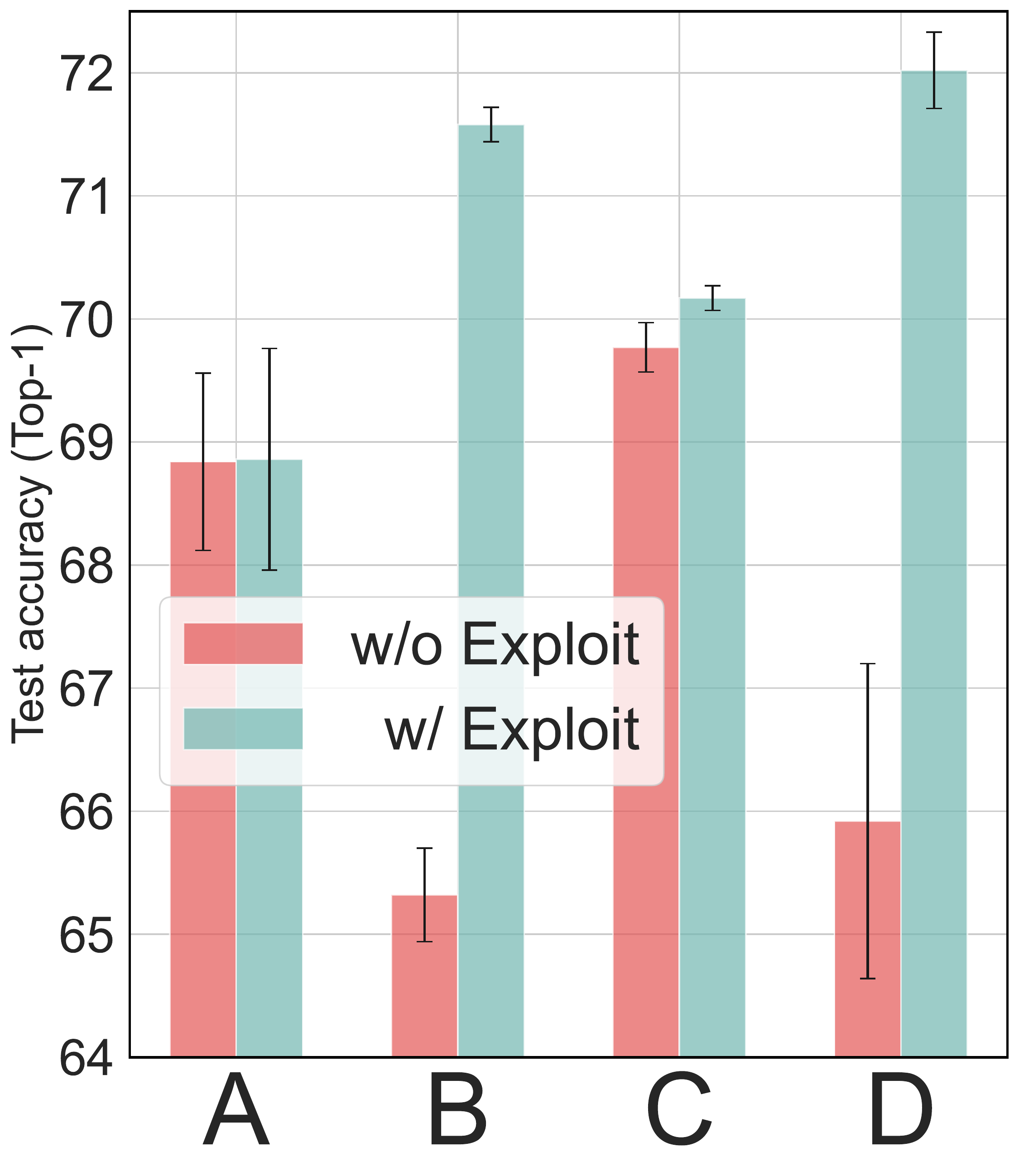}
\vspace{-1.em}
\caption{\textbf{Effect of exploitation} in dynamic model pruning.}
\label{fig:after_exploit}
% \vspace{-1em}
\end{wrapfigure}

In contrast, \textit{well-explored} models (\textsc{Config B, D}) achieve a substantial performance improvement for both the last and best scores.
The training curve (Figure~\ref{fig:comp_bc_exploit}, right) also highlights the effectiveness of exploitation on the model performance; after the target pruning ratio is reached (225 epochs), the well-explored \textsc{Config D} outperforms \textsc{Config C} when boosted by additional exploitation.
Figure~\ref{fig:after_exploit} also illustrates the impact of exploitation by presenting the last score of the pruning process~\cite{kim2021dynamic}. In these results, the models with extensive exploration (\textsc{Config B, D}) exhibit lower final accuracy (and higher deviation) before exploitation; however, their performance (and instability) are greatly improved when further exploitation is introduced.
On the other hand, the models with insufficient exploration (\textsc{Config A, C}) attain only marginal improvement, even though they display higher final scores before exploitation.
A series of experiments reveal that exploitation is essential in the dynamic pruning methodology, and it showcases that our approach is optimally designed to balance exploration and exploitation, thereby enhancing performance and robustness.

\section{Conclusion}

In this work, we have proposed a Magnitude Attention-based Dynamic Model Pruning (MAP) method, which considers element-wise magnitudes in computational paths, unlike conventional pruning methods that treat weights equally in the forward and backward paths, similar to dense model training. MAP consists of exploration and exploitation phases, where the exploitation phase has shown to be highly effective for dynamic pruning methods that undergo exploration.

\textbf{Limitations.\;\;} Our experiments have been primarily conducted on image classification datasets, and we have not yet explored their application to other tasks, such as natural language processing or real-world scenarios. These areas represent avenues for future work. Nevertheless, we believe that we have provided insights into the attention aspect of pruning and the relationship between exploration and exploitation. We believe that our findings will be valuable for future research in this area.

\textbf{Broader impact.\;\;}
Pruning methods have been extensively studied to eliminate redundant parameters and search for sparse networks. However, most existing pruning techniques overlook element-wise weight attention, which considers the importance of weights during the training process. In this work, we introduce a novel approach that leverages weight attention in both the forward and backward paths. The impact of this research extends to various domains that rely on deep neural networks, such as computer vision, natural language processing, and robotics. By enabling more efficient and accurate model pruning, our method offers benefits such as improved resource utilization, reduced energy consumption, and faster inference times. These advancements are particularly valuable in resource-constrained environments, including edge devices, real-time systems, or even Green AI systems more broadly, as computational efficiency is critical in these scenarios.

\clearpage
\input{appendix}

\bibliographystyle{plain}
\bibliography{neurips.bib}

\end{document}

%% file: appendix.tex
\newpage
\appendix
% \begin{center}
%     \textbf{\Large{Supplementary Materials of \\``Magnitude Attention-based Dynamic Pruning"}}
% \end{center}
% \textbf{\Large{Appendix}}

\section{Implementation Details}
For the CIFAR-10/100 datasets~\cite{krizhevsky2009learning}, we train all the models for 300 epochs, with an initial learning rate of 0.2 and an L2 weight decay of $1e-4$. The learning rate is decreased by a factor of 10 at the 150 and 225 epochs, with the latter being the designated epoch for pruning.
In case we use the exploitation phase (Section~\ref{exploitation}), we freeze the sparse structure after 250 epochs.
We use a batch size of 128 (with a single P40 GPU) and set the mask update frequency ($F$) to 16, followed by DPF~\cite{Lin2020Dynamic}.

With regard to ImageNet, we train the model for 100 epochs, incorporating a 10-epoch warm-up phase with an initial learning rate of 0.256.
The learning rate is controlled by a cosine LR schedule~\cite{loshchilov2016sgdr}.
The mask is updated at every iteration ($F=1$). We implement an L1 weight decay of $2e-5$. All experiments are conducted using SGD with a Nesterov momentum of 0.875. The model is trained using four NVIDIA A100 GPUs, achieving a batch size of 1024.
For the exploitation phase, we use the same settings as the CIFAR-10/100 training case.

\section{Pruning Configurations}
Table~\ref{table:a_2_2} formally demonstrates how the forward and backward propagations are performed (as shown in Table~\ref{table:config}).
In particular, we further introduce \textsc{Config D w/o FA} setup (\textsc{Config D} without forward attention), in which the proposed attention mechanism is exclusively applied in the backward path while employing a binary mask for the forward. This additional configuration allows us to understand the criticality of applying attention in both paths by comparing to  \textsc{Config D}.
% A comparative analysis with our proposed method (\textsc{Config E}) enables a better comprehension of the benefits and implications of embedding attention mechanisms in both forward and backward paths.

\begin{table}[h]
    \caption{\textbf{Detailed pruning configurations in the internal study.} We describe the mathematical representations of each pruning configuration as shown in Table~\ref{table:config}. Notably, we include a new configuration, \textsc{Config D w/o FA}, in which the proposed attention is only applied in the backward path, excluding forward attention present in \textsc{Config D}.}
    \centering
    \setlength{\tabcolsep}{1.em}
    \begin{tabular}{c|l|ll}
    \toprule
    Config & Forward path & Backward path \\
    \midrule\midrule
    A & $\overline{w}^{i}_j=m^{(i,P_c)}_{j}\cdot{w}^{i}_j$ & $w_j^{i+1} = w_j^i - \eta \frac{\partial \mathcal{L}}{\partial \overline{w}_j^i} \frac{\partial \overline{w}_j^i}{\partial w_j^i}$ \\
    \midrule
    B & $\overline{w}^{i}_j=m^{(i,P_c)}_{j}\cdot{w}^{i}_j$ & $w_j^{i+1} = w_j^i - \eta \frac{\partial \mathcal{L}}{\partial \overline{w}_j^i}$ \\
    \midrule
    C & $\overline{w}^{i}_j=m^{(i,P_c)}_{j}\cdot{w}^{i}_j$ & $w_j^{i+1} = \begin{cases}
    w_j^i - \eta \frac{\partial \mathcal{L}}{\partial \overline{w}_j^i}, & \text{if } w_j^i \text{ is sparse} \\
    w_j^i - \eta (1-p_c)^z \frac{\partial \mathcal{L}}{\partial \overline{w}_j^i}, & \text{if } w_j^i \text{ is non-sparse}
    \end{cases}$ \\
    \midrule
    \textsc{D w/o FA} & $\overline{w}^{i}_j=m^{(i,P_c)}_{j}\cdot{w}^{i}_j$ & $w_j^{i+1} = \begin{cases}
    w_j^i - \eta \cdot \frac{\partial \mathcal{L}}{\partial \overline{w}^{i}_j}\cdot a^i_j, & \text{if } w_j^i \text{ is sparse} \\
    w_j^i - \eta (1-p_c)^z \frac{\partial \mathcal{L}}{\partial \overline{w}_j^i}, & \text{if } w_j^i \text{ is non-sparse}
    \end{cases}$ \\
    \midrule
    D (ours) & $\widetilde{w}^{i}_j=a^{i}_{j}\cdot\overline{w}^{i}_j$ & $w_j^{i+1} = \begin{cases}
    w_j^i - \eta \cdot \frac{\partial \mathcal{L}}{\partial \widetilde{w}^{i}_j}\cdot a^i_j, & \text{if } w_j^i \text{ is sparse} \\
    w_j^i - \eta (1-p_c)^z \frac{\partial \mathcal{L}}{\partial \widetilde{w}_j^i}, & \text{if } w_j^i \text{ is non-sparse}
    \end{cases}$ \\
    \bottomrule
    \end{tabular}
    \label{table:a_2_2}
\end{table}

\textbf{\textsc{Config A}} employs a binary mask for both the forward and backward paths, adhering to the approach introduced by Zhu \textit{et al.}~\cite{zhu2017prune}. In this configuration, the binary mask is utilized to selectively prune weights in both directions.

\textbf{\textsc{Config B}} bears resemblance to \textsc{Config A} in the forward path as it employs a sparse binary mask. However, for the backward path, it adopts the straight-through estimator (STE)~\cite{bengio2013estimating} that enables equivalent updates of both sparse and non-sparse models. This configuration takes its cues from DPF as proposed by Lin \textit{et al.}~\cite{Lin2020Dynamic}.

\textbf{\textsc{Config C}} also adopts a binary mask in the forward path, however, this utilizes the pruning retention ratio $(1-p_c)^z$ weighting scheme for the non-sparse model in the backward path. With this weighting, as the pruning ratio increases, the concentration of the non-sparse model progressively decreases, indicating that sparse components are prioritized during the learning process. We have observed the effectiveness of the weighting scheme that differentially updates sparse and non-sparse components based on the current pruning ratio $p_c$. Consequently, we have integrated the weighting scheme that updates according to the pruning retention ratio into our final methodology (\textsc{Config D}).

\textbf{\textsc{Config D (ours)}} is the introduced MAP method. We leverage the proposed magnitude attention mechanism as defined by the equation in Eq.~\ref{attention} for both forward and backward paths and it allows us to guide the overall model update in the direction steered by important layers.
This gives precedence to critical layers and enhances the efficacy of sparse model updates. Furthermore, the non-sparse model is updated utilizing the $(1-p_c)^z$ weighting scheme (as same as \textsc{Config C}), fostering proactive exploration during the training process.
% The distinction between \textsc{Config D} and \textsc{E} underscores our demonstration that the consideration of the attention algorithm is necessary in both directions.

\textbf{\textsc{Config D w/o FA}} applies a binary mask in the forward path, contrary to \textsc{Config D}, which integrates the attention-based mask. By comparing this setting to \textsc{Config D} (ours), we demonstrate the necessity of the proposed magnitude attention in both directions.

\section{Model Analysis}

In addition to the experimental analysis presented in Section~\ref{exploitation} (in the main text), we conduct further experiments to increase the understanding of the dynamic pruning including the proposed MAP.
We hope that the supplemental results provide more insights into the effectiveness and benefits of MAP.

\begin{table}[t]
\caption{\textbf{Comparison of pruning configurations on CIFAR-10.} We report test accuracy using ResNet-20/32/56 (shortly R20/R32/R56) backbones. All the results include the last and best epochs' scores to assess the training stability (smaller deviation indicates higher stability).}
\small
\setlength{\tabcolsep}{0.55em}
\centering
    \begin{tabular}{ccccccccc}
    \toprule
                              &                       & Dense              & Exploit? & A           & B           & C           & \textsc{D w/o FA}           & D           \\
    \midrule\midrule
    \multirow{4}[2]{*}{R20} & \multirow{2}{*}{Last} & \multirow{2}{*}{\perf{92.22}{0.15}} & \textcolor{red}{\ding{55}} & \perf{90.79}{0.07}  & \perf{87.84}{1.91} & \perf{\textbf{90.90}}{0.05}  & \perf{88.07}{0.44} & \perf{88.32}{1.19} \\
                              &                       &                   & \textcolor{PineGreen}{\ding{51}}  & \perf{90.81}{0.01} & \perf{91.35}{0.17} & \perf{91.02}{0.16} & \perf{91.51}{0.11} & \perf{\textbf{92.12}}{0.06} \\           
                              \cmidrule{2-9}
                              & \multirow{2}{*}{Best}  & \multirow{2}{*}{\perf{92.36}{0.10}} & \textcolor{red}{\ding{55}} & \perf{91.03}{0.06}  & \perf{90.79}{0.07} & \perf{\textbf{91.13}}{0.12}  & \perf{91.03}{0.02} & \perf{91.36}{0.12} \\
                              &                        &                    & \textcolor{PineGreen}{\ding{51}}  & \perf{91.10}{0.03} & \perf{91.47}{0.22} & \perf{91.35}{0.20} & \perf{91.64}{0.13} & \perf{\textbf{92.19}}{0.04} \\   
    \midrule
    \multirow{4}[2]{*}{R32} & \multirow{2}{*}{Last}  & \multirow{2}{*}{\perf{93.16}{0.01}} & \textcolor{red}{\ding{55}} & \perf{91.93}{0.21}  & \perf{91.01}{0.33} & \perf{\textbf{92.24}}{0.15} & \perf{90.77}{0.27} & \perf{91.79}{0.24} \\  
                              &                        &                    & \textcolor{PineGreen}{\ding{51}}  & \perf{92.02}{0.20} & \perf{92.42}{0.29} & \perf{92.05}{0.27} & \perf{92.79}{0.13} & \perf{\textbf{92.97}}{0.11} \\
                              \cmidrule{2-9}
                              & \multirow{2}{*}{Best}  & \multirow{2}{*}{\perf{93.22}{0.07}} & \textcolor{red}{\ding{55}} & \perf{92.14}{0.13}  & \perf{92.30}{0.16} & \perf{92.50}{0.16}  & \perf{92.54}{0.10} & \perf{\textbf{92.73}}{0.01} \\
                              &                        &                    & \textcolor{PineGreen}{\ding{51}}  & \perf{92.19}{0.05}  & \perf{92.56}{0.28} & \perf{92.33}{0.35} & \perf{92.86}{0.16} & \perf{\textbf{93.18}}{0.03} \\     
    \midrule              
    \multirow{4}[2]{*}{R56} & \multirow{2}{*}{Last}  & \multirow{2}{*}{\perf{93.86}{0.11}}  & \textcolor{red}{\ding{55}} & \perf{92.82}{0.37}  & \perf{92.62}{0.12} & \perf{\textbf{93.29}}{0.19} & \perf{92.78}{0.03} & \perf{93.18}{0.12} \\  
                              &                        &                    & \textcolor{PineGreen}{\ding{51}}  & \perf{93.19}{0.17} & \perf{93.31}{0.35} & \perf{93.31}{0.08} & \perf{93.52}{0.02} & \perf{\textbf{93.84}}{0.08} \\
                              \cmidrule{2-9}
                              & \multirow{2}{*}{Best}  &  \multirow{2}{*}{{\perf{94.03}{0.16}}}        &     \textcolor{red}{\ding{55}}      & \perf{93.00}{0.53}  & \perf{93.16}{0.09} & \perf{93.51}{0.23}  & \perf{93.18}{0.13} & \perf{\textbf{93.54}}{0.07} \\
                              &                        &                    & \textcolor{PineGreen}{\ding{51}}  & \perf{93.34}{0.20} & \perf{93.44}{0.30} & \perf{93.59}{0.05} & \perf{93.61}{0.04} & \perf{\textbf{94.02}}{0.04} \\     
    \bottomrule
    \end{tabular}
\label{table:config_c10}
\end{table}

\subsection{Comparison of Pruning Configurations}
Here, we report the full comparison results of each pruning configuration as depicted in Table~\ref{table:a_2_2}. In this evaluation, we use ResNet-20/32/56~\cite{he2016deep} (shortly R20/R32/R56) as the backbone network and we validate all the methods on the CIFAR-10/100~\cite{krizhevsky2009learning} datasets. We experiment with five different setups (A, B, C, D, and \textsc{D w/o FA}) recording both the last and best scores for each.

As shown in Table~\ref{table:config_c10} and \ref{table:config_c100} (and also discussed in Section~\ref{learning}), \textsc{Config A} and \textsc{C} primarily focus on training the sparse model (exploitation), resulting in stable learning and a small accuracy difference between the last and best scores~\cite{kim2021dynamic}. However, due to limited exploration, the further exploitation effect is not prominent in these configurations.

On the other hand, \textsc{Config B, D}, and \textsc{D w/o FA} effectively train both the non-sparse and sparse models, facilitating superior exploration. Consequently, when reaching the exploitation phase of fixing the structure of the sparse network in the later stages of training, a significant performance improvement is demonstrated. Nevertheless, we can also observe the huge deviation between the last and best accuracy due to the training instability derived by the higher exploration.

The proposed pruning method, \textsc{Config D} (MAP) outperforms previous dynamic pruning models such as DPF~\cite{Lin2020Dynamic} (i.e. \textsc{Config B}) in terms of performance.
This could be attributed to the magnitude attention mechanism's ability to update critical layers more efficiently. While the pruning method focuses on training important layers rather than sub-layers (similar to \textsc{Config C}), our approach allows for continuous updates to both sparse and non-sparse networks instead of discrete updates.
This continuous updating facilitates enhanced exploration, leading to remarkable exploration performance and effective exploitation. Consequently, magnitude attention plays a crucial role in enabling effective learning, surpassing other models in terms of training efficacy and overall performance.

To further investigate the effect of the magnitude attention, we compare \textsc{Config D} and \textsc{D w/o FA}, which applies the attention mechanism only backward propagation.
The comparison demonstrates the efficacy of introducing attention to both training paths.
By incorporating attention in the backward path, we motivate more frequent updates to important layers' weights. Utilizing attention in the forward path during the training process directs the model updates towards the guidance of significant layers, leading to performance improvement.

\begin{table}[t]
\caption{\textbf{Comparison of pruning configurations on CIFAR-100.} We report test accuracy using ResNet-20/32/56 (shortly R20/R32/R56) backbones. All the results include the last and best epochs' scores to assess the training stability (smaller deviation indicates higher stability).}
\centering
\small
\setlength{\tabcolsep}{0.55em}
\begin{tabular}{ccccccccc}
    \toprule
                              &                        & Dense    & Exploit?          & A           & B           & C           & \textsc{D w/o FA}          & D           \\
    \midrule\midrule
    \multirow{4}[2]{*}{R20} & \multirow{2}{*}{Last}        & \multirow{2}{*}{\perf{67.86}{0.41}} & \textcolor{red}{\ding{55}} & \perf{\textbf{65.21}}{0.36}  & \perf{52.55}{2.58} & \perf{64.99}{0.36}  & \perf{54.01}{0.65} & \perf{57.62}{1.42} \\
                              &                            &         & \textcolor{PineGreen}{\ding{51}}           & \perf{65.61}{0.13}  & \perf{65.63}{0.27} & \perf{66.13}{0.40} & \perf{65.89}{0.25} & \perf{\textbf{67.04}}{0.21} \\
                              \cmidrule{2-9}
                              & \multirow{2}{*}{Best}     &  \multirow{2}{*}{\perf{68.64}{0.22}}  & \textcolor{red}{\ding{55}} & \perf{65.67}{0.36}  & \perf{63.79}{0.20} & \perf{\textbf{65.86}}{0.28} & \perf{63.95}{0.30} & \perf{65.13}{0.51} \\  
                              &                              &           & \textcolor{PineGreen}{\ding{51}}         & \perf{65.99}{0.11}  & \perf{65.79}{0.26} & \perf{66.46}{0.27} & \perf{65.92}{0.32} & \perf{\textbf{67.11}}{0.02} \\  
    \midrule
    \multirow{4}[2]{*}{R32} & \multirow{2}{*}{Last}      & \multirow{2}{*}{\perf{69.89}{0.21}} & \textcolor{red}{\ding{55}} & \perf{67.34}{0.38}  & \perf{56.87}{1.83} & \perf{\textbf{67.94}}{0.27}  & \perf{57.45}{2.80} & \perf{60.24}{2.58} \\
                              &                            &       & \textcolor{PineGreen}{\ding{51}}             & \perf{67.45}{0.19}  & \perf{69.72}{0.18} & \perf{68.44}{0.09} & \perf{69.74}{0.14} & \perf{\textbf{70.17}}{0.12} \\
                              \cmidrule{2-9}
                              & \multirow{2}{*}{Best}     & \multirow{2}{*}{\perf{70.60}{0.07}} & \textcolor{red}{\ding{55}} & \perf{67.91}{0.42}  & \perf{67.88}{0.48} & \perf{\textbf{68.75}}{0.23} & \perf{67.70}{0.34} & \perf{68.58}{0.52} \\ 
                              &                            &    & \textcolor{PineGreen}{\ding{51}}            & \perf{68.00}{0.16}  & \perf{69.69}{0.10} & \perf{68.79}{0.05} & \perf{69.95}{0.19} & \perf{\textbf{70.42}}{0.14} \\  
    \midrule    
    \multirow{4}[2]{*}{R56} & \multirow{2}{*}{Last}      & \multirow{2}{*}{\perf{72.01}{0.31}} & \textcolor{red}{\ding{55}} & \perf{68.84}{0.72}  & \perf{65.32}{0.38} & \perf{\textbf{69.77}}{0.20}  & \perf{64.29}{1.33} & \perf{65.92}{1.28} \\
                              &                           &             & \textcolor{PineGreen}{\ding{51}}       & \perf{68.86}{0.90}  & \perf{71.58}{0.14} & \perf{70.17}{0.10} & \perf{71.62}{0.05} & \perf{\textbf{72.02}}{0.31} \\
                              \cmidrule{2-9}
                              & \multirow{2}{*}{Best}   & \multirow{2}{*}{\perf{72.25}{0.12}}  & \textcolor{red}{\ding{55}} & \perf{69.33}{0.84}  & \perf{70.44}{0.15} & \perf{70.60}{0.13} & \perf{70.89}{0.34} & \perf{\textbf{70.93}}{0.31} \\  
                              &                          &          & \textcolor{PineGreen}{\ding{51}}          & \perf{69.43}{0.84}  & \perf{71.64}{0.11} & \perf{70.79}{0.06} & \perf{71.78}{0.04} & \perf{\textbf{72.24}}{0.13} \\  
    \bottomrule
    \end{tabular}
    \label{table:config_c100}
\end{table}

\begin{table}[htbp]
  \centering
  \caption{\textbf{Test accuracy on various attention strengths $z$.} We present the accuracy on CIFAR-10 with ResNet-20/32 (R20/R32) backbones. The result demonstrates that higher attention strength $z$ leads to improved performance of the pruned models.}
  % \caption{\textbf{Training Results based on Attention Strength for ResNet-20/32 on CIFAR-10.} The table presents the variation in test accuracy based on different attention strengths ($z$) for the pruning retention ratio ($(1-p_c)^z$). The results indicate that higher attention strength leads to improved performance of the models.}
  \small
  \begin{tabular}{c|cccccccc}
    \toprule
    $z=$ & 0.4 & 0.6 & 0.8 & 1.0 & 1.2 & 1.4 & 1.6 \\
    \midrule\midrule
    R20 & \perf{88.20}{0.09} & \perf{88.72}{0.06} & \perf{89.03}{0.50} & \perf{89.30}{0.41} & \perf{90.00}{0.11} & \perf{90.27}{0.12} & \perf{90.60}{0.23} \\
    \midrule
    R32 & \perf{91.12}{0.28} & \perf{91.29}{0.14} & \perf{91.82}{0.13} & \perf{91.80}{0.13} & \perf{92.11}{0.10} & \perf{91.93}{0.48} & \perf{92.47}{0.28} \\
    \bottomrule
  \end{tabular}
  \label{tab:accuracy_scores}
\end{table}

\begin{figure}[h]
    \centering
    \includegraphics[width=1\linewidth]{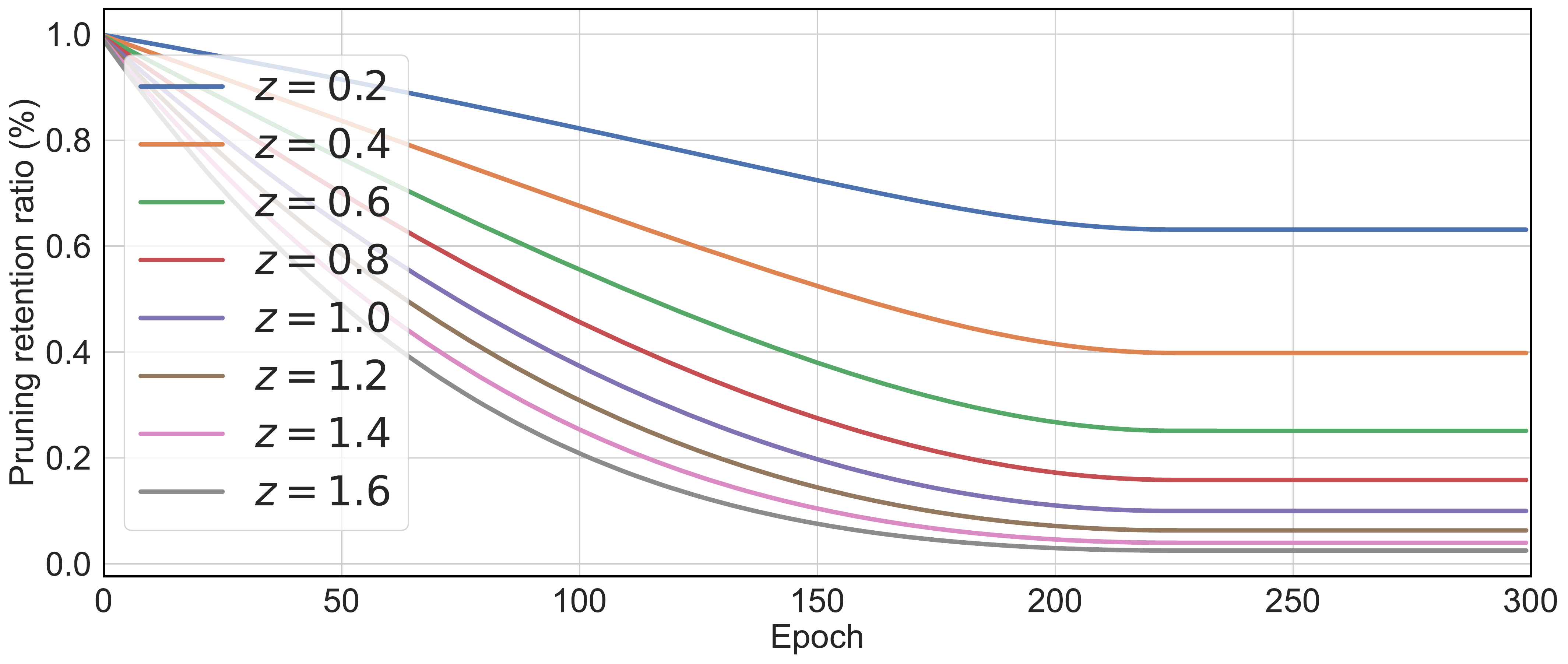}
    \caption{\textbf{Pruning retention ratio.} The pruning retention ratio, represented as $(1-p_c)^z$, is influenced by the current pruning ratio ($p_c$) and decreases as the attention strength $z$ increases.}
    \label{fig:suppl_retention}
\end{figure}

\begin{figure}[t]
\centering
\includegraphics[width=1.0\linewidth]{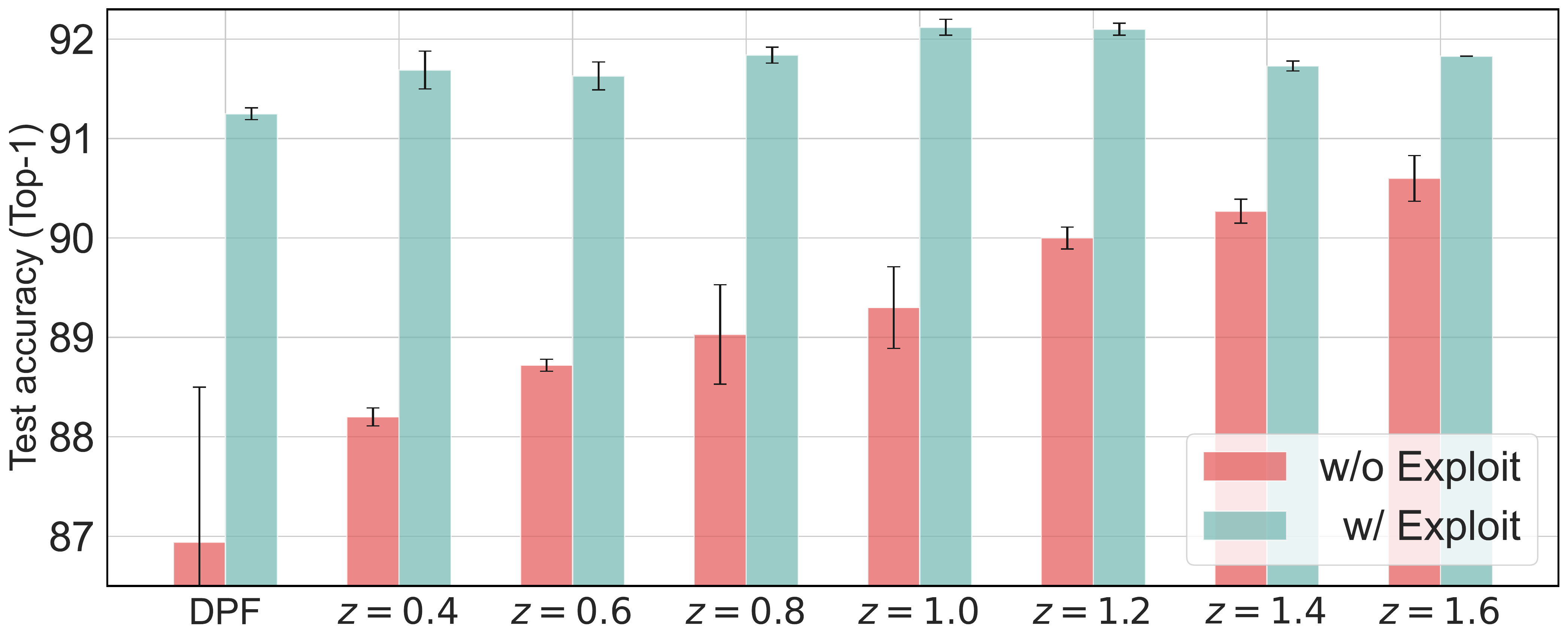}
\caption{\textbf{Effect of attention strength w/ and w/o exploitation.} We show the result on CIFAR-10 using ResNet-20 model. All the scores are obtained in the last epoch, following Kim \textit{et al.}~\cite{kim2021dynamic}.}
\label{fig:att_str2}
\end{figure}

\subsection{Impact of Attention Strength}
\label{sec:imp_att_str}
In our pruning method, the attention strength is influenced by the parameter $z$ in the formulation $(1-p)^z$ of Eq.(\ref{attention}).
When $z$ is small, the value of $(1-p)^z$ increases, resulting in relatively weaker attention. Conversely, when $z$ is large, the value of $(1-p)^z$ decreases, leading to a stronger and more pronounced attention effect (Figure~\ref{fig:suppl_retention}). 
Through various experiences across multiple $z$ values, we consistently observed that as $z$ continues to grow, the attention exhibits heightened sensitivity to weight importance, thereby fostering a discerning focus on preserving and updating pivotal connections. Moreover, we found that increasing $z$ also corresponded to improved performance outcomes (Figure~\ref{fig:att_str2}). Thus, these results serve as strong evidence that our attention-based methodology excels in identifying high-performing sparse networks and effectively enhancing their performance, surpassing the efficacy of conventional dynamic pruning approaches.

\subsection{Impact of Using Both Attention and Exploitation}

To investigate the advantages of the proposed attention-based pruning with the combined effects of exploitation, we conduct a series of experimental analyses.
The proposed attention methodology achieves effective training of sparse networks by employing differential weight updates based on the magnitude importance of weights.
Moreover, continuous updates are ensured for both sparse and non-sparse networks, leading to improved exploration performance (Section~\ref{main/5.3.1.mag_att}).
By virtue of our methodology designed to facilitate robust exploration, the structural integrity of sparse networks is preserved, leading to significant performance gains when exploitation is employed (Section~\ref{analysis_exploitation}).

\begin{figure}[h]
    \centering
    \includegraphics[width=1\linewidth]{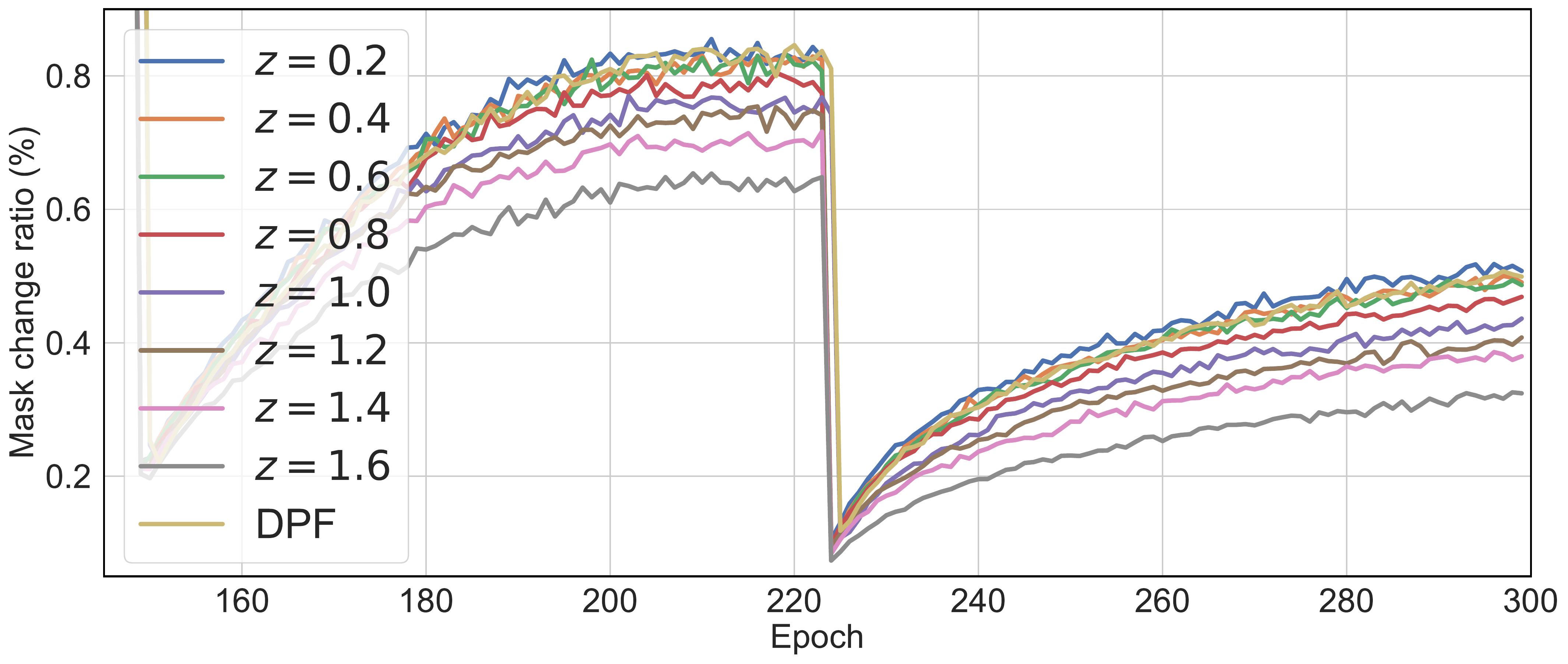}
    \caption{\textbf{Pruning mask change ratio} on various attention strengths $z$.}
    \label{fig:suppl_mask_change}
\end{figure}

Nevertheless, finding a sweet spot between exploration and exploitation could be challenging, owing to their inherent trade-off relationship.
% 여기 rephrase해서 넣으려 했는데 아래 글이랑 스토리라인이 반대라서 이상해요
% For instance, elevating the strength of attention led to enhanced identification of well-performing sparse networks (higher performance before the exploitation phase). However, higher $z$ diminishes exploration capability, which induces a similar performance across various $z$ after further exploitation  (Figure~\ref{fig:att_str2}).
Remarkably, we observed that larger $z$ values reduce the transition from non-sparse to sparse networks (Figure~\ref{fig:suppl_mask_change}).
In other words, strengthening the attention effect can be seen as prioritizing exploitation rather than exploring the overall model, which may diminish the effectiveness of fixing the structure of a well-explored sparse network during the latter stages of training. Consequently, when both attention and exploitation are utilized, irrespective of attention strength, similar performance outcomes are observed (Figure~\ref{fig:att_str2}). Therefore, all experiments were conducted with a fixed $z$ value of 1.0 for simplicity.

% 일단 말이 이상하고.. 굳이 안넣어도 될거같아요
% These findings underscore the intricacy of concurrently optimizing attention strength and exploitation, as achieving an ideal balance remains elusive. Nevertheless, our experiments provide invaluable insights into the interplay between attention and exploitation within the context of our pruning methodology. By effectively balancing both exploration and exploitation strategies, we were able to achieve remarkable performance outcomes.

\subsection{Design Choice of Attention Mask}
The proper designing of an attention mask, as shown in Eq.(\ref{attention}), involves aligning the weights of the sparse model layer based on their magnitudes and then normalizing these values within the range of $(1-p_c)^z$ to $1.0$. The normalization process can be performed with two alternative approaches: linear and non-linear normalization.

In \textbf{linear normalization}, the attention mask values are scaled to fit within the desired range of $(1-p_c)^z$ to $1.0$, while preserving their original order of weight magnitude. This approach utilizes linear interpolation to transform the non-linear weights into a linear range of $(1-p_c)^z$ to $1.0$.

On the other hand, \textbf{non-linear normalization} takes into account the relative sizes of the weight magnitudes to create an attention mask within the range of $(1-p_c)^z$ to $1.0$. This approach offers more flexibility in adjusting the distribution of the attention mask. By employing non-linear mapping, specific weights can be amplified or attenuated, allowing for precise fine-tuning of the attention distribution across layers according to the specific requirements of the model.

\begin{table}[t]
\centering
\caption{\textbf{Test Accuracy of linear and non-linear schemes.} We present the test accuracy of both linear and non-linear schemes of the attention design for ResNet-20/32/56 trained on CIFAR-10.}
\footnotesize
\setlength{\tabcolsep}{0.55em}
\begin{tabular}{cccccccc}
\toprule
\multirow{2}[2]{*}{Attention} &  \multirow{2}[2]{*}{Exploit?}  & \multicolumn{2}{c}{ResNet-20} & \multicolumn{2}{c}{ResNet-32} & \multicolumn{2}{c}{ResNet-56} \\
\cmidrule{3-8}
&     & Last          & Best         & Last          & Best         & Last          & Best         \\
\midrule\midrule
\multirow{2}{*}{Linear} &  \textcolor{red}{\ding{55}} & \perf{69.70}{2.62}   & \perf{86.48}{0.44}  & \perf{76.06}{6.57}   & \perf{88.97}{0.39}  & \perf{76.30}{3.37}        & \perf{90.93}{0.14}       \\
       & \textcolor{PineGreen}{\ding{51}}  & \perf{90.29}{0.14}        & \perf{90.38}{0.15}       & \perf{92.22}{0.10}        & \perf{92.26}{0.02}       & \perf{91.66}{0.34}        & \perf{91.87}{0.09}       \\
    \midrule
    Non-Linear   &  \textcolor{red}{\ding{55}} & \perf{88.32}{1.19}        & \perf{91.36}{0.12}       & \perf{91.79}{0.24}   & \perf{92.73}{0.01}       & \perf{93.18}{0.12}   & \perf{93.54}{0.07}  \\
       (Ours)   & \textcolor{PineGreen}{\ding{51}}  & \perf{92.12}{0.06}        & \perf{92.19}{0.04}  & \perf{92.97}{0.11}        & \perf{93.18}{0.03}  & \perf{93.84}{0.08}        & \perf{94.02}{0.04}  \\
    \bottomrule
\end{tabular}
\label{tab:att_design_result}
\end{table}

\begin{figure}[h]
\centering
\includegraphics[width=\linewidth]{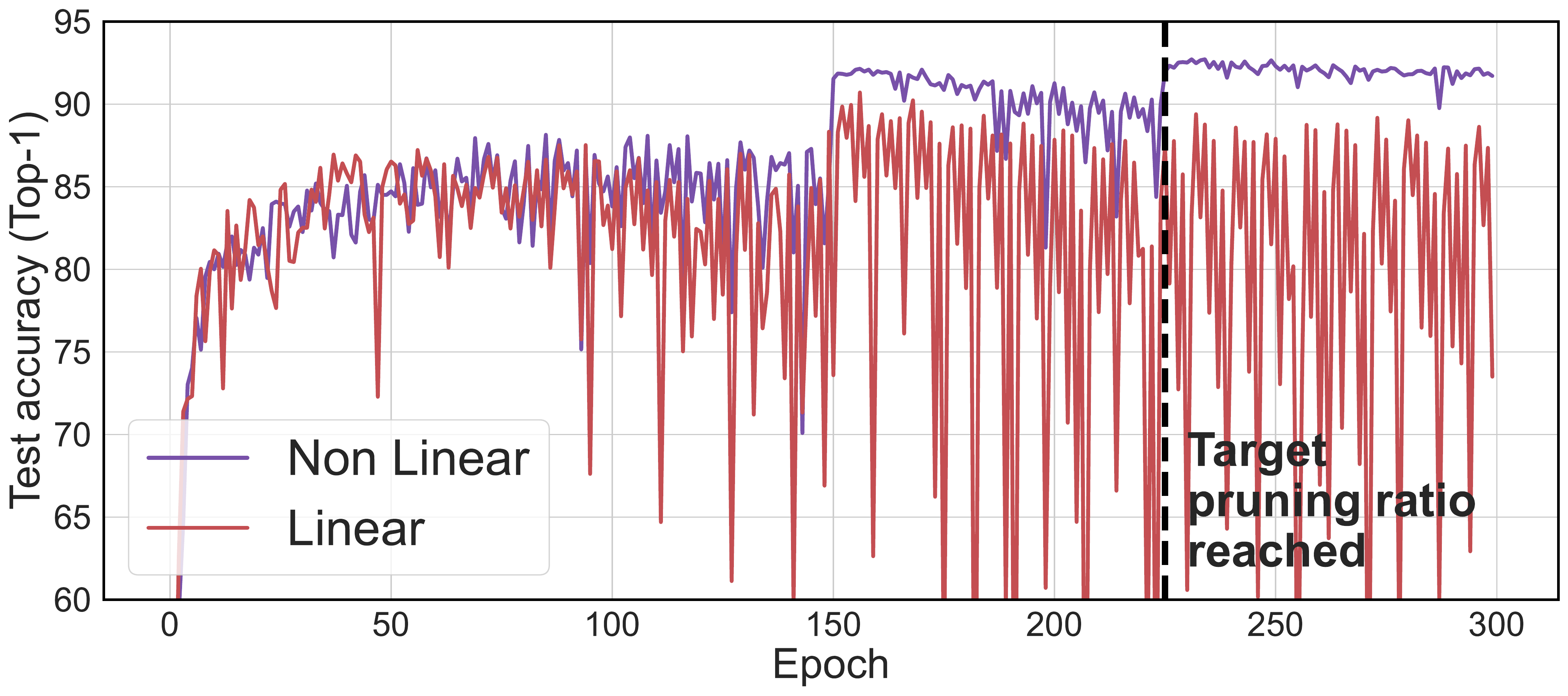}
\caption{\textbf{Training of linear and non-Linear schemes of attention design choices.}}
\label{fig:att_design_training}
\end{figure}

We compare the linear and non-linear normalization approach on our attention mask calculations, and the results are presented in Table~\ref{tab:att_design_result}.
Linear normalization solely relies on sorting the weights in order of magnitude to generate the attention mask, without considering the relative differences between weights that undergo fine changes during the learning process. Consequently, this approach may lead to a situation where weights with relatively small magnitudes receive strong attention in the ranked layer, potentially impacting learning performance negatively.
In contrast, non-linear normalization, which allows precise adjustments over weight magnitude distribution provides appropriate attention to the weights by well-referring its magnitude.
Indeed, as depicted in Figure~\ref{fig:att_design_training}, we observe that learning stability improved when attention was naturally assigned based on the weight magnitudes (i.e. non-linear normalization), rather than forcibly applying linear mapping. In addition, it is also verified that the performance significantly improved when the attention mask is proportionate to the weight magnitude (Table~\ref{tab:att_design_result}).

%% file: main.bbl
\begin{thebibliography}{10}

\bibitem{alizadeh2022prospect}
Milad Alizadeh, Shyam~A Tailor, Luisa~M Zintgraf, Joost van Amersfoort,
  Sebastian Farquhar, Nicholas~Donald Lane, and Yarin Gal.
\newblock Prospect pruning: Finding trainable weights at initialization using
  meta-gradients.
\newblock {\em arXiv preprint arXiv:2202.08132}, 2022.

\bibitem{DBLP:conf/iclr/BellecK0L18}
Guillaume Bellec, David Kappel, Wolfgang Maass, and Robert Legenstein.
\newblock Deep rewiring: Training very sparse deep networks.
\newblock In {\em 6th International Conference on Learning Representations,
  {ICLR} 2018, Vancouver, BC, Canada, April 30 - May 3, 2018, Conference Track
  Proceedings}. OpenReview.net, 2018.

\bibitem{bengio2013estimating}
Yoshua Bengio, Nicholas L{\'e}onard, and Aaron Courville.
\newblock Estimating or propagating gradients through stochastic neurons for
  conditional computation.
\newblock {\em arXiv preprint arXiv:1308.3432}, 2013.

\bibitem{8578988}
Miguel~A. Carreira-Perpinan and Yerlan Idelbayev.
\newblock "learning-compression" algorithms for neural net pruning.
\newblock In {\em 2018 IEEE/CVF Conference on Computer Vision and Pattern
  Recognition}, pages 8532--8541, 2018.

\bibitem{chen2023unified}
Yanqi Chen, Zhengyu Ma, Wei Fang, Xiawu Zheng, Zhaofei Yu, and Yonghong Tian.
\newblock A unified framework for soft threshold pruning.
\newblock {\em International Conference on Learning Representations}, 2023.

\bibitem{chen2022state}
Yanqi Chen, Zhaofei Yu, Wei Fang, Zhengyu Ma, Tiejun Huang, and Yonghong Tian.
\newblock State transition of dendritic spines improves learning of sparse
  spiking neural networks.
\newblock In {\em International Conference on Machine Learning}, pages
  3701--3715. PMLR, 2022.

\bibitem{deng2009imagenet}
Jia Deng, Wei Dong, Richard Socher, Li-Jia Li, Kai Li, and Li~Fei-Fei.
\newblock Imagenet: A large-scale hierarchical image database.
\newblock In {\em 2009 IEEE conference on computer vision and pattern
  recognition}, pages 248--255. Ieee, 2009.

\bibitem{dettmers2019sparse}
Tim Dettmers and Luke Zettlemoyer.
\newblock Sparse networks from scratch: Faster training without losing
  performance.
\newblock {\em arXiv preprint arXiv:1907.04840}, 2019.

\bibitem{NEURIPS2019_f34185c4}
Xiaohan Ding, guiguang ding, Xiangxin Zhou, Yuchen Guo, Jungong Han, and
  Ji~Liu.
\newblock Global sparse momentum sgd for pruning very deep neural networks.
\newblock In H.~Wallach, H.~Larochelle, A.~Beygelzimer, F.~d\textquotesingle
  Alch\'{e}-Buc, E.~Fox, and R.~Garnett, editors, {\em Advances in Neural
  Information Processing Systems}, volume~32. Curran Associates, Inc., 2019.

\bibitem{frankle2018lottery}
Jonathan Frankle and Michael Carbin.
\newblock The lottery ticket hypothesis: Finding sparse, trainable neural
  networks.
\newblock {\em arXiv preprint arXiv:1803.03635}, 2018.

\bibitem{guo2016dynamic}
Yiwen Guo, Anbang Yao, and Yurong Chen.
\newblock Dynamic network surgery for efficient dnns.
\newblock {\em Advances in neural information processing systems}, 29, 2016.

\bibitem{gupta2022complexity}
Manas Gupta, Efe Camci, Vishandi~Rudy Keneta, Abhishek Vaidyanathan, Ritwik
  Kanodia, Chuan-Sheng Foo, Wu~Min, and Lin Jie.
\newblock Is complexity required for neural network pruning? a case study on
  global magnitude pruning.
\newblock {\em arXiv preprint arXiv:2209.14624}, 2022.

\bibitem{han2015learning}
Song Han, Jeff Pool, John Tran, and William Dally.
\newblock Learning both weights and connections for efficient neural network.
\newblock {\em Advances in neural information processing systems}, 28, 2015.

\bibitem{hayou2020robust}
Soufiane Hayou, Jean-Francois Ton, Arnaud Doucet, and Yee~Whye Teh.
\newblock Robust pruning at initialization.
\newblock {\em International Conference on Learning Representations}, 2021.

\bibitem{he2016deep}
Kaiming He, Xiangyu Zhang, Shaoqing Ren, and Jian Sun.
\newblock Deep residual learning for image recognition.
\newblock In {\em Proceedings of the IEEE conference on computer vision and
  pattern recognition}, pages 770--778, 2016.

\bibitem{ioffe2015batch}
Sergey Ioffe and Christian Szegedy.
\newblock Batch normalization: Accelerating deep network training by reducing
  internal covariate shift.
\newblock In {\em International conference on machine learning}, pages
  448--456. pmlr, 2015.

\bibitem{kim2021dynamic}
Jangho Kim, Jayeon Yoo, Yeji Song, KiYoon Yoo, and Nojun Kwak.
\newblock Dynamic collective intelligence learning: Finding efficient sparse
  model via refined gradients for pruned weights.
\newblock {\em arXiv preprint arXiv:2109.04660}, 2021.

\bibitem{kim2020position}
Jangho Kim, KiYoon Yoo, and Nojun Kwak.
\newblock Position-based scaled gradient for model quantization and pruning.
\newblock {\em Advances in neural information processing systems},
  33:20415--20426, 2020.

\bibitem{krizhevsky2009learning}
Alex Krizhevsky, Geoffrey Hinton, et~al.
\newblock Learning multiple layers of features from tiny images.
\newblock 2009.

\bibitem{kusupati2020soft}
Aditya Kusupati, Vivek Ramanujan, Raghav Somani, Mitchell Wortsman, Prateek
  Jain, Sham Kakade, and Ali Farhadi.
\newblock Soft threshold weight reparameterization for learnable sparsity.
\newblock In {\em International Conference on Machine Learning}, pages
  5544--5555. PMLR, 2020.

\bibitem{le2021network}
Duong~H Le and Binh-Son Hua.
\newblock Network pruning that matters: A case study on retraining variants.
\newblock {\em International Conference on Learning Representations}, 2021.

\bibitem{lecun1989optimal}
Yann LeCun, John Denker, and Sara Solla.
\newblock Optimal brain damage.
\newblock {\em Advances in neural information processing systems}, 2, 1989.

\bibitem{lee2018snip}
Namhoon Lee, Thalaiyasingam Ajanthan, and Philip~HS Torr.
\newblock Snip: Single-shot network pruning based on connection sensitivity.
\newblock {\em arXiv preprint arXiv:1810.02340}, 2018.

\bibitem{Lin2020Dynamic}
Tao Lin, Sebastian~U. Stich, Luis Barba, Daniil Dmitriev, and Martin Jaggi.
\newblock Dynamic model pruning with feedback.
\newblock In {\em International Conference on Learning Representations}, 2020.

\bibitem{liu2020dynamic}
Junjie Liu, Zhe Xu, Runbin Shi, R.~Cheung, and Hayden Kwok-Hay So.
\newblock Dynamic sparse training: Find efficient sparse network from scratch
  with trainable masked layers.
\newblock {\em International Conference On Learning Representations}, 2020.

\bibitem{liu2021sparse}
Shiwei Liu, Tianlong Chen, Xiaohan Chen, Zahra Atashgahi, Lu~Yin, Huanyu Kou,
  Li~Shen, Mykola Pechenizkiy, Zhangyang Wang, and Decebal~Constantin Mocanu.
\newblock Sparse training via boosting pruning plasticity with
  neuroregeneration.
\newblock {\em Advances in Neural Information Processing Systems},
  34:9908--9922, 2021.

\bibitem{loshchilov2016sgdr}
Ilya Loshchilov and Frank Hutter.
\newblock Sgdr: Stochastic gradient descent with warm restarts.
\newblock {\em arXiv preprint arXiv: Arxiv-1608.03983}, 2016.

\bibitem{Constantin2018scalable}
Decebal~Constantin Mocanu, Elena Mocanu, Peter Stone, Phuong~H. Nguyen,
  Madeleine Gibescu, and Antonio Liotta.
\newblock Scalable training of artificial neural networks with adaptive sparse
  connectivity inspired by network science.
\newblock {\em Nature Communications}, 2018.

\bibitem{mostafa2019parameter}
Hesham Mostafa and Xin Wang.
\newblock Parameter efficient training of deep convolutional neural networks by
  dynamic sparse reparameterization.
\newblock In {\em International Conference on Machine Learning}, pages
  4646--4655. PMLR, 2019.

\bibitem{renda2020comparing}
Alex Renda, Jonathan Frankle, and Michael Carbin.
\newblock Comparing rewinding and fine-tuning in neural network pruning.
\newblock {\em International Conference on Learning Representations}, 2020.

\bibitem{singh2020woodfisher}
Sidak~Pal Singh and Dan Alistarh.
\newblock Woodfisher: Efficient second-order approximation for neural network
  compression.
\newblock {\em Advances in Neural Information Processing Systems},
  33:18098--18109, 2020.

\bibitem{srinivas2022cyclical}
Suraj Srinivas, Andrey Kuzmin, Markus Nagel, Mart van Baalen, Andrii Skliar,
  and Tijmen Blankevoort.
\newblock Cyclical pruning for sparse neural networks.
\newblock In {\em Proceedings of the IEEE/CVF Conference on Computer Vision and
  Pattern Recognition}, pages 2762--2771, 2022.

\bibitem{Wang2020Picking}
Chaoqi Wang, Guodong Zhang, and Roger Grosse.
\newblock Picking winning tickets before training by preserving gradient flow.
\newblock In {\em International Conference on Learning Representations}, 2020.

\bibitem{wortsman2019discovering}
Mitchell Wortsman, Ali Farhadi, and Mohammad Rastegari.
\newblock Discovering neural wirings.
\newblock {\em Advances in Neural Information Processing Systems}, 32, 2019.

\bibitem{zhang2021understanding}
Chiyuan Zhang, Samy Bengio, Moritz Hardt, Benjamin Recht, and Oriol Vinyals.
\newblock Understanding deep learning (still) requires rethinking
  generalization.
\newblock {\em Communications of the ACM}, 64(3):107--115, 2021.

\bibitem{zhang2018learning}
Dejiao Zhang, Haozhu Wang, Mario Figueiredo, and Laura Balzano.
\newblock Learning to share: Simultaneous parameter tying and sparsification in
  deep learning.
\newblock In {\em International Conference on Learning Representations}, 2018.

\bibitem{zhou2021effective}
Xiao Zhou, Weizhong Zhang, Hang Xu, and Tong Zhang.
\newblock Effective sparsification of neural networks with global sparsity
  constraint.
\newblock In {\em Proceedings of the IEEE/CVF Conference on Computer Vision and
  Pattern Recognition}, pages 3599--3608, 2021.

\bibitem{zhu2017prune}
Michael Zhu and Suyog Gupta.
\newblock To prune, or not to prune: exploring the efficacy of pruning for
  model compression.
\newblock {\em arXiv preprint arXiv:1710.01878}, 2017.

\bibitem{zimmer2023how}
Max Zimmer, Christoph Spiegel, and Sebastian Pokutta.
\newblock How i learned to stop worrying and love retraining.
\newblock In {\em The Eleventh International Conference on Learning
  Representations}, 2023.

\end{thebibliography}
